\documentclass{article}

\usepackage{microtype}
\usepackage{graphicx}
\usepackage{subcaption}
\usepackage{booktabs} 
\usepackage{pifont}  
\usepackage{minitoc} 

\usepackage{hyperref}
\newcommand{\revise}[1]{\textcolor{black}{#1}}
\usepackage{siunitx}

\newcommand{\res}[2]{#1\textsubscript{\textpm #2}}
\usepackage[preprint]{icml2026}
\usepackage{multirow} 

\usepackage{amsmath}
\usepackage{amssymb}
\usepackage{mathtools}
\usepackage{amsthm}

\usepackage{xcolor}
\usepackage{adjustbox}
\usepackage{makecell}
\usepackage{pifont}   

\newcommand{\yes}{\textcolor{green}{\ding{51}}} 
\newcommand{\no}{\textcolor{red}{\ding{55}}}   
\definecolor{darkgreen}{RGB}{0,100,0}
\definecolor{darkred}{RGB}{139,0,0}

\usepackage[capitalize,noabbrev]{cleveref}
\usepackage{tcolorbox}

\newtcolorbox{promptbox}[1][]{
  colback=gray!5!white,   
  colframe=gray!75!black, 
  title={\textbf{Prompt}}, 
  fonttitle=\bfseries,
  left=5pt, right=5pt, top=5pt, bottom=5pt,
  boxrule=1pt,
  sharp corners=south, 
  #1
}

\newcommand{\code}[1]{\colorbox{gray!15}{\texttt{#1}}}
\theoremstyle{plain}

\theoremstyle{definition}

\theoremstyle{remark}

\icmltitlerunning{Med-REFL: Enhancing Medical Complex Reasoning via Fine-grained Self-Correction}

\begin{document}

\doparttoc 
\faketableofcontents 

\twocolumn[
  \icmltitle{Med-REFL: Enhancing Complex Medical Reasoning via Fine-grained Self-Correction}

  \icmlsetsymbol{equal}{*}

  \begin{icmlauthorlist}
    \icmlauthor{Zongxian Yang}{inst1}
    \icmlauthor{Jiayu Qian}{inst1}
    \icmlauthor{Zegao Peng}{inst1}
    \icmlauthor{Haoyu Zhang}{inst1}
    \icmlauthor{Yu-An Huang}{inst2}
    \icmlauthor{KC Tan}{inst3}
    \icmlauthor{Zhi-An Huang}{inst1}
  \end{icmlauthorlist}

  \icmlaffiliation{inst1}{City University of Hong Kong (Dongguan), China}
  \icmlaffiliation{inst2}{Northwestern Polytechnical University, China}
  \icmlaffiliation{inst3}{The Hong Kong Polytechnic University, China}

  \icmlcorrespondingauthor{Zhi-An Huang}{huang.za@cityu-dg.edu.cn}


  \vskip 0.3in
]

\printAffiliationsAndNotice{}

\begin{abstract}
Large reasoning models (LRMs) excel in domains like mathematics where intermediate reasoning is straightforward to verify, but struggle to self-correct in medicine fields where evaluating intermediate reasoning is cumbersome and expensive. This verification bottleneck hinders the development of reliable AI reasoners for high-stakes application. Here we propose \textbf{Med-REFL}, a novel framework that learns fine-grained reflection without human labels or model distillation. Med-REFL introduces a deterministic structural assessment of the reasoning space to automatically generate preference data for reflection. By globally evaluating all explored reasoning paths in a tree-of-thoughts, our method quantifies the value of reflection actions, enabling the automated construction of direct preference optimization pairs. This trains the model to recognize and amend its own reasoning fallacies.  
Extensive experiments show that Med-REFL robust gains on both in-domain and out-of-distribution tasks across multiple models, and it also yields an average improvement of 3.59\% over the state-of-the-art models on seven benchmarks.
Crucially, targeted ablations prove its success \revise{generalizes to other domains such as logical reasoning and successfully mitigates the fake reflection phenomenon in LRMs.} 
Ultimately, our framework provides a feasible solution to the verification bottleneck in medical chain-of-thoughts, paving the way for more reliable AI reasoners in high-stakes domains like medicine.
\end{abstract}

\section{Introduction}

\begin{figure}[t]
    \vspace{-10pt}
    \centering
    \includegraphics[width=0.8\linewidth]{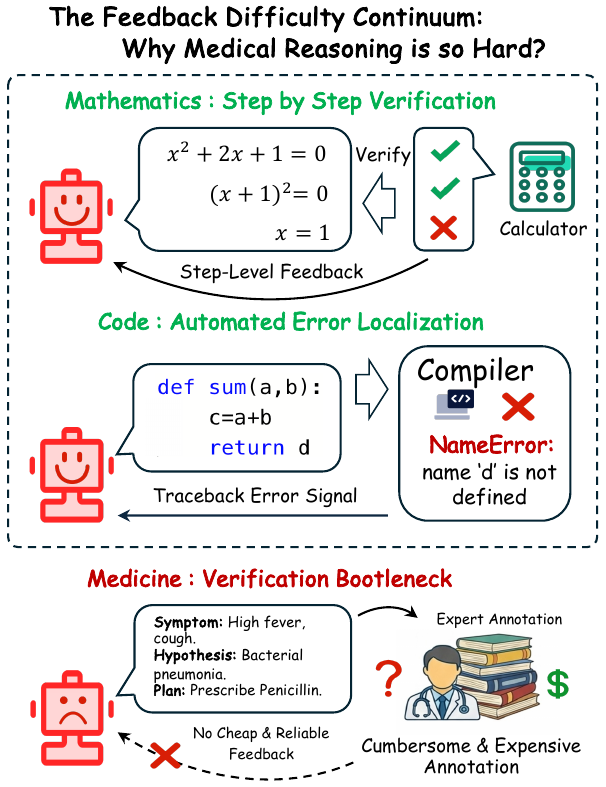}
    \caption{The verification bottleneck across different reasoning domains. Compared to Mathematics (where steps are verified by calculators) and Code (where compilers provide automated error localization), Medicine suffers from a verification bottleneck. }
    \label{fig:Introduce}
    \vspace{-10pt}
\end{figure}

Recent advancements in large reasoning models (LRMs) like Deepseek-R1 ~\citep{deepseekai2025deepseekr1incentivizingreasoningcapability} have demonstrated remarkable progress, especially in code and math fields. A core character is the spontaneous emergence of self-reflection during the reasoning process~\citep{plaat2024reasoninglargelanguagemodels}. 
Inspired by this, researchers began to train medical LRMs wjth supervised fine-tuning (SFT)~\citep{dong2023abilities} with distilled long chain of thought (CoT) ~\citep{huatuogpto1} or GRPO~\citep{liu2025distillationpushinglimitsmedical}.
However, these attempts often yield suboptimal performance and, more crucially, fail to observe effective reflection behaviors. 
As shown in Figure ~\ref{fig:Introduce}, the fundamental cause preventing the effective migration of this paradigm from mathematics and code to the medical domain lies in the significant differences in verification accessibility. 
Unlike the former, which can leverage low-cost, automated tools such as calculators or compilers to provide precise error signals for model refinement, medical reasoning lacks such convenient and inexpensive `compilers'~\citep{hui2024qwen2, li2025system}. 
Instead, validating clinical reasoning typically necessitates human experts annotation or cross-referencing against unstructured medical database, which makes the feedback loop prohibitively expensive and cumbersome~\citep{Med333, medreason, qiu2025quantifyingreasoningabilitiesllms}. 
Without effective correction signals to guide the transition from erroneous to correct states, models remain unable to internalize the self-correction mechanism essential for complex medical reasoning~\citep{zhang2024learningcheckunleashingpotentials, kamoi2024can, xi2024traininglargelanguagemodels}.

Despite these challenges, previous attempts to enhance medical reasoning have made valuable progress. The first paradigm, relying on SFT with data distilled from costly teacher models or on outcome-based reinforcement learning (RL)~\citep{cheng2023mrrl,dai2024safe}, is inherently result-oriented. These methods primarily rely on final answers as selection or reward signals, paying insufficient attention to intermediate processes~\citep{xi2024traininglargelanguagemodels}. This makes distillation methods prone to learning flawed logic that coincidentally produces correct answers, while outcome-based RL is vulnerable to reward hacking, both failing to guarantee robust reasoning~\citep{miao2024inform,bai2022training}. The second paradigm, which leverages process reward models (PRMs)~\citep{li2025process}, provides more granular stepwise feedback. However, this approach often necessitates training a separate, demanding reward model and creates a complex, multi-stage pipeline~\citep{zhang2025lessonsdevelopingprocessreward}. More critically, this paradigm is fundamentally designed to identify and reward the most correct path, which does not effectively teach the model the essential skill of how to reflect on and recover from its own diverse errors~\citep{NEURIPS2023_91edff07}.

To overcome these limitations, we introduce \textbf{Med-REFL}, a novel framework that learns fine-grained reflection within a resource-efficient system. Our approach begins with a deterministic structural assessment of the reasoning space, using a tree-of-thoughts (ToT)~\citep{long2023large,yao2023tree} to map out diverse reasoning pathways. By aggregating outcomes over the entire tree, we derive a step value for each intermediate state to mitigate the `reward hacking' problem. To enable the model to learn self-correction, we further introduce the action value, which utilizes the tree's global information to evaluate reflection actions. This metric measures how much these actions improve the entire problem-solving trajectory, thereby guiding the model's reflective reasoning learning. Action value guides the construction of high-quality preference data for direct preference optimization (DPO)~\citep{rafailov2023direct}, enabling model to distinguish between invalid and effective reflection.

Our extensive experiments validate this approach from multiple perspectives. \textbf{First}, as described in Table~\ref{tab:comprehensive_results}, Med-REFL demonstrates remarkable versatility by enhancing models in different post-training paradigms. It yields the most substantial improvements on reason-heavy models, with an average gain of +3.59\% at Huatuo-o1-8B~\citep{huatuogpto1}, pushing already state-of-the-art (SoTA) models to new performance heights. \textbf{Second}, the framework shows robust generalization, delivering a significant +3.88\% average gain on the in-distribution MedQA-USMLE~\citep{jin2021disease}) benchmark while maintaining strong performance across a range of challenging out-of-distribution(OoD) datasets like GPQA (+3.89\%)~\citep{rein2024gpqa}. \textbf{Third}, our method is effective not only for medical tasks but also for logic and real-world clinical problems, which showcases its strong generalization ability.
Most importantly, targeted ablations reveal that Med-REFL can mitigate the fake reflection phenomenon in LRMs~\citep{kang2025first,kamoi2024can} and substantially enhance the robustness of reasoning.

\begin{table}[t!]
\centering
\small
\caption{Comparison of different LLM reasoning strategies based on their core attributes. Med-RELF addresses the limitations of previous approaches. Expert refers to closed-source teacher model or human experts.}
\resizebox{\linewidth}{!}{
\begin{tabular}{l c c c c c c} 
\toprule
\textbf{Method} & 
\textbf{\makecell{Medical \\ Specific}} & 
\textbf{\makecell{Reflect \\ Ability}} & 
\textbf{\makecell{Process \\ Supervision}} & 
\textbf{\makecell{\revise{Expert} \\ Free}} & 
\textbf{\makecell{\revise{PRM} \\ Free}} & 
\textbf{\makecell{RAG \\ Free}} \\
\midrule
Huatuo-o1~\citep{huatuogpto1} & \yes & \yes & \no & \no & \yes & \yes \\
MedReason~\citep{medreason} & \yes & \no & \yes & \no & \yes & \no \\
O1-Journy~\citep{o1-part2} & \no & \yes & \no & \no & \yes & \yes \\
AlphaMed~\citep{liu2025distillationpushinglimitsmedical} & \yes & \no & \no & \yes & \yes & \yes \\
Med-PRM~\citep{medprm} & \yes & \no & \yes & \no & \no & \no \\
Med$^3$~\citep{Med333} & \yes & \no & \yes & \yes & \no & \yes \\ 
Med-REFL(ours) & \yes & \yes & \yes & \yes & \yes & \yes \\
\bottomrule
\end{tabular}
}
\label{tab:strategy_comparison}
\end{table}

Our main contributions are as follows:
\begin{enumerate}
    \vspace{-10pt}
    \item We propose Med-REFL a novel framework to overcome the verification bottleneck of in medical reasoning. To our knowledge, our approach is the first to successfully enhance medical LRM through focusing on improving reflection.
    \vspace{-4pt}
    \item We introduce a deterministic, structure-based evaluation that yields an experimentally-validated learning signal, enabling the automatic generation of fine-grained reflection data and overcoming the limitations of both outcome-oriented learning and PRMs.
    \vspace{-4pt}
    \item We demonstrate through extensive experiments that Med-REFL consistently improves performance across various models and benchmarks, and is capable of training new SoTA models. Via targeted ablations, \revise{we further prove that this success stems from internalized self-correction, which effectively mitigates the pervasive fake reflection in LRMs.}
\end{enumerate}

\section{Med-REFL Overview}
\label{headings}

\begin{figure*}
    \centering
    \includegraphics[width=1\linewidth]{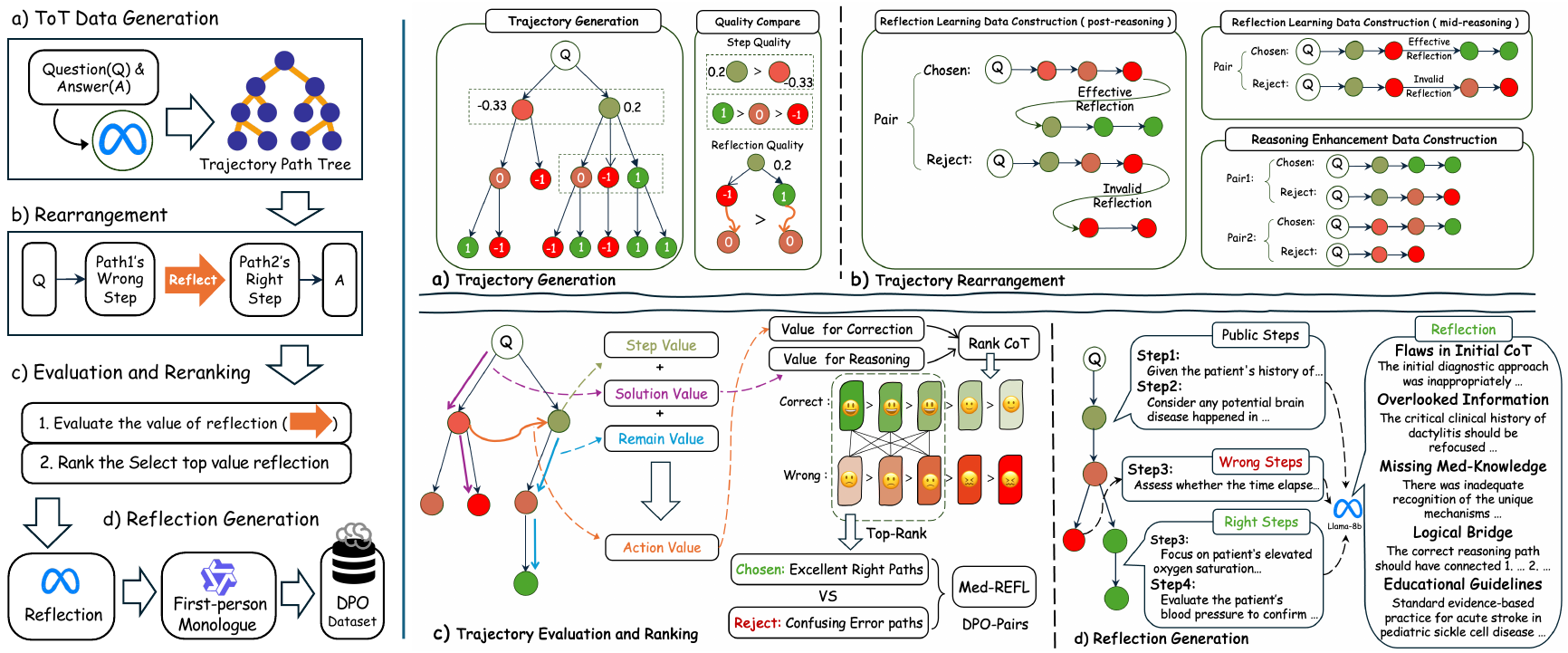}
    \caption{The figure presents the pipeline in two views: the left outlines the high-level workflow, while the right details the specific mechanisms for each stage. Med-REFL consists of four stages: a) illustrates how trajectories rolls out ; b) illustrates the target data pair consåtruction; c) illustrates how action value calculated and how to rank reasoning path; d) An example showcases the fine-grained self-correction mechanism where initial reasoning flaws are identified and recovered during medical question answering.}
    \label{fig:overview}
\end{figure*}

Our methodology is designed to teach the model a instilled reflection capability. The pipeline, illustrated in Figure~\ref{fig:overview}, comprises four main stages: (1) generating a diverse map of reasoning trajectories ToT approach; (2) establishing a deterministic, structure-based method to quantitatively evaluate the quality of any intermediate step and any reflection action; (3) leveraging these metrics to automatically construct two types of preference pairs—a primary set for `Reflection Learning' and a supplementary set for `Reasoning Enhancement'; and (4) fine-tuning the model using DPO to instill a robust self-correction capability. The comparison of different reasoning enhancement strategies can be found in Table~\ref{tab:strategy_comparison}.

\subsection{Trajectory Generation and Evaluation}
\label{ssec:trajectory}

To generate diverse reasoning trajectories, each node represents an intermediate reasoning state, and a path from root to leaf forms a complete CoT trajectory. ToT (details in Appendix~\ref{subapdx:tot}) explore multiple reasoning directions at each step to generate diverse and high-quality trajectories.

To evaluate trajectory quality, we compare the final answer of its each path to the ground truth, assigning `+1' to steps in correct paths and `-1' to steps in incorrect paths. For a step $s$, we define its quality as:
\begin{equation}
    v_{\text{step}}(s) = \frac{N_{s,\text{correct}} - N_{s,\text{incorrect}}}{N_s},
\end{equation}
where $N_s$ is the number of paths in the subtree rooted at $s$, and $N_{s,\text{correct}}$ and $N_{s,\text{incorrect}}$ are the counts of correct and incorrect paths, respectively. For a root-to-leaf trajectory $P = (s_1, \dots, s_L)$ of length $L$, we compute its solution quality:
\begin{equation}
\label{eq:solution_value}
    v_{\text{sol}}(P) = \frac{1}{L} \sum_{i=1}^{L} v_{\text{step}}(s_i).
\end{equation}
Similarly, for a partial node-to-leaf trajectory starting at step $j$, the remaining quality is:
\begin{equation}
    v_{\text{rem}}(P') = \frac{1}{L-j+1} \sum_{i=j}^{L} v_{\text{step}}(s_i).
\end{equation}
To assess reflection actions, we define the action value for transitioning from node $s$ to $s'$ via action $a$:
\begin{multline}
\label{eq:action_value}
v_{\text{act}}(a_{(s,s')}) = \lambda_1 ( v_{\text{step}}(s)-v_{\text{step}}(s')) \\
+ \lambda_2 (v_{\text{sol}}(s)-v_{\text{sol}}(s')) + \lambda_3 (v_{\text{rem}}(s)-v_{\text{rem}}(s'))
\end{multline}
where $\lambda_1, \lambda_2, \lambda_3$ are tunable weights.

\subsection{Preference Pair Construction for Reflection Learning}
\label{ssec:reflection}

To train large language model (LLM) for error correction, we construct DPO pairs that favor trajectories where reflection corrects errors over those where reflection fails. For a medical question $q$ with ground truth $y$ and its candidate CoT  trajectory $P$, an error locator and a reflector are driven by pre-trained LLM $\pi$. This process uses two LLM-based tools:
\begin{itemize}
    \item \textbf{Error Locator ($E_\pi$):} Identifies the first incorrect step $s_{\text{err}}$ in an erroneous CoT trajectory $P_{\text{err}}$ with it's candidate node set, $S_{\text{cand}}$, formed by the $k$ lowest $v_{\text{step}}$ score nodes:  
    \begin{equation}
        s_{\text{err}} = E_\pi(q, y, S_{\text{cand}}(k),P_{\text{err}}).
        \label{eq:serro}
    \end{equation}
    \item \textbf{Reflector ($C_\pi$):} Generates reflection text $R$ explaining the transition from erroneous steps $S_{\text{orig}}$ to corrected steps $S_{\text{new}}$, given common prefix steps $S_{\text{pub}}$:
    \begin{equation}
        R = C_\pi(q, S_{\text{pub}}, S_{\text{orig}}, S_{\text{new}}).
    \end{equation}

\end{itemize}

Starting from the parent node of $s_{\text{err}}$, the model explores alternative reasoning branches, classifying them as correct or incorrect based on their final answers. Next, we construct algorithms to artificially simulate two scenarios of reflection in reasoning:
\begin{itemize}
    \item Firstly, we simulate scenarios where the model, upon encountering a challenge or \textbf{detecting an error during the reasoning process (mid-reasoning)} to correct a single erroneous step $s_{\text{err}}$: $R = C_\pi(q, S_{\text{pub}}, s_{\text{err}}, S_{\text{new}})$, with the trajectory: $P_{\text{refl}} = S_{\text{pub}} \rightarrow s_{\text{err}} \rightarrow R \rightarrow S_{\text{new}}$.
    \item In the second, we simulate the \textbf{reflection occurs post-reasoning}, regarding as a review after the entire erroneous trajectory from $s_{\text{err}}$ to the trajectory’s end ($S_{\text{orig}}$): $R' = C_\pi(q, S_{\text{pub}}, S_{\text{orig}}, S_{\text{new}})$, with the trajectory: $P'_{\text{refl}} = S_{\text{pub}} \rightarrow S_{\text{orig}} \rightarrow R' \rightarrow S_{\text{new}}$.
    \item Finally, trajectories are ranked by $v_{\text{act}}$ (Equation~\ref{eq:action_value}), and we select a top-ranked effective reflection trajectory ($P_{\text{chosen}}$) and a top-ranked invalid reflection trajectory (reflect but lead to another wrong step) ($P_{\text{rejected}}$) as seed trajectories to construct the first-person reasoning trace $r$:
\begin{equation}
\label{eq:rMP}
\begin{cases}
r_{chosen} = M_\pi(P_{chosen}), \\
r_{rejected} = M_\pi(P_{rejected}),
\end{cases}
\end{equation}
where $M_\pi$ is a thinker to transform $P_{\text(ref)}$ or $P'_{\text{refl}}$ into a verbalized reasoning trace $r$ by reformulating the explicit steps in P into natural language thinking. 
\end{itemize}

So the high-quality reasoning trace $r_{chosen}$ and error-prone reasoning $r_{rejected}$ could be generated by $M_\pi$ from $P_{\text{chosen}}$ and $P_{\text{rejected}}$ to form the DPO pair ($r_{\text{chosen}}, r_{\text{rejected}}$).

\subsection{Preference Pair Construction for Reasoning Enhancement}
\label{ssec:reasoning}
To maintain broad reasoning proficiency, we supplement the reflection data with general-purpose reasoning pairs by comparing high-quality correct CoT trajectories with plausible but incorrect ones. Like reflection enhancement pairs, trajectories are ranked by $V_{\text{sol}}$, and similar to Equation~\ref{eq:rMP}, we select a correct high-quality trajectory ($P'_\text{chosen}$) and a plausible incorrect trajectory ($P'_{\text{rejected}}$) to generate the reasoning trace $r'$ using $M_{\pi}$ resulting in the DPO pair ($r'_{\text{chosen}}, r'_{\text{rejected}}$). 

\begin{table*}[!t]
\centering
\scriptsize
\setlength{\tabcolsep}{1pt}
\caption{A comprehensive evaluation of models with different post-training methods. The table shows original accuracy (\%, orig.) with standard deviations in subscripts and the performance gain from Med-REFL (w/ ours) across various in-distribution and OoD medical benchmarks. }
\resizebox{\linewidth}{!}{%
\begin{tabular}{l l cc cc cc cc cc cc cc c}
\toprule
\multicolumn{2}{c}{\textbf{Post-Training Method}} & \multicolumn{4}{c}{\textbf{Mix}} & \multicolumn{8}{c}{\textbf{SFT}} & \multicolumn{2}{c}{\textbf{GRPO}} & \\
\cmidrule(lr){3-6} \cmidrule(lr){7-14} \cmidrule(lr){15-16}
\multicolumn{2}{c}{\textbf{Model Type}} & \multicolumn{4}{c}{\textbf{Instruction-Tuned}} & \multicolumn{4}{c}{\textbf{Reason-Heavy}} & \multicolumn{4}{c}{\textbf{Knowledge-Heavy}} & \multicolumn{2}{c}{\textbf{Pure-RL}} & \\
\cmidrule(lr){3-6} \cmidrule(lr){7-10} \cmidrule(lr){11-14} \cmidrule(lr){15-16} 
\textbf{  Domain} & \textbf{  Dataset} & \multicolumn{2}{c}{LLaMa3.1-8B} & \multicolumn{2}{c}{Qwen2.5-7B} & \multicolumn{2}{c}{Huatuo-o1-8B} & \multicolumn{2}{c}{\makecell{Deepseek-\\Distill-8B}} & \multicolumn{2}{c}{\makecell{MedReason\\-8B}} & \multicolumn{2}{c}{\makecell{UltraMedical\\3.1-8B}} & \multicolumn{2}{c}{\makecell{AlphaMed\\-8B}} & \textbf{Avg.} \\
\cmidrule(lr){3-4} \cmidrule(lr){5-6} \cmidrule(lr){7-8} \cmidrule(lr){9-10} \cmidrule(lr){11-12} \cmidrule(lr){13-14} \cmidrule(lr){15-16}
& & orig. & w/ ours & orig. & w/ ours & orig. & w/ ours & orig. & w/ ours & orig. & w/ ours & orig. & w/ ours & orig. & w/ ours & \\
\midrule

\multirow{2}{*}{\textbf{In-Distribution}} & MedQA-U (5 op) 
& \res{59.92}{1.34} & \res{65.88}{1.21} 
& \res{57.11}{0.80} & \res{60.05}{1.17}  
& \res{69.59}{0.70} & \textbf{\res{73.74}{0.70}}  
& \res{48.85}{2.17} & \res{55.07}{0.55}   
& \res{66.27}{1.80} & \res{70.50}{1.11}  
& \res{71.34}{0.93} & \res{73.43}{0.88}   
& \res{67.79}{1.13} & \res{69.34}{0.77}   
& \\
&  & & \textcolor{darkgreen}{+5.96} & & \textcolor{darkgreen}{+2.94} & & \textcolor{darkgreen}{+4.15} & & \textbf{\textcolor{darkgreen}{+6.22}} & & \textcolor{darkgreen}{+4.23} & & \textcolor{darkgreen}{+2.09} & & \textcolor{darkgreen}{+1.55} & \textcolor{darkgreen}{+3.88} \\
\midrule

\multirow{2}{*}{\makecell[l]{\textbf{OoD}}} 
& MedMCQA 
& \res{57.61}{0.32} & \res{59.40}{0.47} 
& \res{54.52}{0.29} & \res{55.84}{0.11} 
& \res{62.13}{0.44} & \textbf{\res{65.00}{0.17}} 
& \res{49.51}{0.41} & \res{50.12}{0.41} 
& \res{58.98}{0.33} & \res{60.00}{1.35} 
& \res{63.30}{0.40} & \res{64.36}{0.23} 
& \res{61.22}{0.58} & \res{62.29}{0.26} 
& \\
&  & & \textcolor{darkgreen}{+1.79} & & \textcolor{darkgreen}{+1.32} & & \textbf{\textcolor{darkgreen}{+2.87}} & & \textcolor{darkgreen}{+0.61} & & \textcolor{darkgreen}{+1.02} & & \textcolor{darkgreen}{+1.06} & & \textcolor{darkgreen}{+1.07} & \textcolor{darkgreen}{+1.39} \\

& GPQA(M) 
& \res{45.16}{1.43} & \res{50.47}{0.55} 
& \res{43.34}{0.62} & \res{45.44}{0.90} 
& \res{50.67}{1.83} & \res{57.01}{1.66} 
& \res{62.43}{1.02} & \textbf{\res{65.28}{0.85}} 
& \res{45.64}{1.14} & \res{50.09}{1.31} 
& \res{45.76}{2.09} & \res{49.42}{1.40} 
& \res{53.40}{2.06} & \res{55.92}{0.98} 
& \\
&  & & \textcolor{darkgreen}{+5.31} & & \textcolor{darkgreen}{+2.10} & & \textbf{\textcolor{darkgreen}{+6.34}} & & \textcolor{darkgreen}{+2.85} & & \textcolor{darkgreen}{+4.45} & & \textcolor{darkgreen}{+3.66} & & \textcolor{darkgreen}{+2.52} & \textcolor{darkgreen}{+3.89} \\

& MedXpert-R 
& \res{14.14}{0.58} & \res{15.66}{0.12} 
& \res{12.45}{0.16} & \res{13.62}{0.20} 
& \res{16.85}{0.22} & \res{18.06}{0.97} 
& \res{11.98}{0.78} & \res{13.62}{0.37} 
& \res{22.31}{0.77} & \res{22.47}{0.29} 
& \res{15.87}{0.24} & \res{16.38}{0.52} 
& \res{21.91}{1.08} & \textbf{\res{22.68}{0.72}} 
& \\
&  & & \textcolor{darkgreen}{+1.52} & & \textcolor{darkgreen}{+1.17} & & \textcolor{darkgreen}{+1.21} & & \textbf{\textcolor{darkgreen}{+1.64}} & & \textcolor{darkgreen}{+0.16} & & \textcolor{darkgreen}{+0.51} & & \textcolor{darkgreen}{+0.77} & \textcolor{darkgreen}{+1.00} \\

& MedXpert-U 
& \res{16.11}{0.82} & \res{18.21}{0.27} 
& \res{10.79}{0.47} & \res{12.91}{0.18} 
& \res{15.31}{1.42} & \res{20.09}{1.14} 
& \res{13.07}{0.86} & \res{15.23}{0.86} 
& \res{22.55}{0.34} & \textbf{\res{25.58}{1.63}} 
& \res{15.68}{1.53} & \res{16.18}{1.56} 
& \res{20.24}{1.08} & \res{20.84}{0.67} 
& \\
&  & & \textcolor{darkgreen}{+2.10} & & \textcolor{darkgreen}{+2.12} & & \textbf{\textcolor{darkgreen}{+4.78}} & & \textcolor{darkgreen}{+2.16} & & \textcolor{darkgreen}{+3.03} & & \textcolor{darkgreen}{+0.50} & & \textcolor{darkgreen}{+0.60} & \textcolor{darkgreen}{+2.18} \\

& MMLU-Pro(M) 
& \res{57.56}{0.97} & \res{61.17}{0.82} 
& \res{61.36}{0.13} & \res{62.71}{0.34} 
& \res{61.87}{0.96} & \res{65.08}{0.99} 
& \res{53.83}{1.02} & \res{56.28}{0.03} 
& \res{59.14}{1.34} & \res{62.80}{0.80} 
& \res{63.06}{0.20} & \res{63.55}{0.87} 
& \res{63.29}{0.67} & \textbf{\res{65.09}{1.09}} 
& \\
&  & & \textcolor{darkgreen}{+3.61} & & \textcolor{darkgreen}{+1.35} & & \textcolor{darkgreen}{+3.21} & & \textcolor{darkgreen}{+2.45} & & \textbf{\textcolor{darkgreen}{+3.66}} & & \textcolor{darkgreen}{+0.49} & & \textcolor{darkgreen}{+1.80} & \textcolor{darkgreen}{+2.37} \\

& PubMedQA  
& \res{76.26}{0.59} & \res{77.80}{0.50} 
& \res{74.12}{0.24} & \res{75.09}{0.40} 
& \res{79.02}{0.73} & \textbf{\res{81.60}{0.28}} 
& \res{69.30}{1.66} & \res{70.71}{1.98} 
& \res{77.17}{0.80} & \res{77.96}{0.46} 
& \res{78.60}{0.24} & \res{78.73}{0.21} 
& \res{78.03}{0.13} & \res{77.74}{0.83} 
& \\
&  & & \textcolor{darkgreen}{+1.54} & & \textcolor{darkgreen}{+0.97} & & \textbf{\textcolor{darkgreen}{+2.58}} & & \textcolor{darkgreen}{+1.41} & & \textcolor{darkgreen}{+0.79} & & \textcolor{darkgreen}{+0.13} & & \textcolor{darkred}{-0.29} & \textcolor{darkgreen}{+1.02} \\
\midrule

\multirow{2}{*}{\textbf{Average}} & Accuracy & 46.68 & 49.80 & 44.81 & 46.52 & 50.78 & \textbf{54.37} & 44.14 & 46.62 & 50.29 & 52.77 & 50.52 & 51.72 & 52.27 & 53.42 & \\
&  & & \textcolor{darkgreen}{+3.12} & & \textcolor{darkgreen}{+1.71} & & \textbf{\textcolor{darkgreen}{+3.59}} & & \textcolor{darkgreen}{+2.48} & & \textcolor{darkgreen}{+2.48} & & \textcolor{darkgreen}{+1.21} & & \textcolor{darkgreen}{+1.15} & \textcolor{darkgreen}{+2.25} \\
\bottomrule
\end{tabular}
}
\vspace{-5px}
\label{tab:comprehensive_results}
\end{table*}

\subsection{Model Fine-Tuning with DPO}
\label{ssec:finetuning}

After the previous three-stage process, we obtain the 33k DPO data: $D^{+}_\text{DPO}=(\{r_{\text{chosen}},{r_{\text{reject}}}\},\{r'_{\text{chosen}},r'_{{\text{reject}}}\})$. Then we fine-tune LLMs with $D^{+}_\text{DPO}$ to prioritize effective reasoning and reflection patterns. This process optimizes the model to favor effective reflection instead of invalid reflection, enabling robust error correction in medical question answering.

\section{Main Experiments}
\label{sec:main_experiments}

\subsection{Experimental Setup}
\paragraph{Benchmark.}
Our primary dataset for developing and evaluating Med-REFL is the MedQA-USMLE (5 options)~\citep{jin2021disease} question-answering task. 
To further assess the generalization ability of models fine-tuned with Med-REFL, we conducted evaluations on several additional diverse medical benchmarks, including MedMCQA (validation set)~\citep{pal2022medmcqa}, PubMedQA (test set)~\citep{jin2019pubmedqa}, GPQA medical subset (GPQA-M) and MMLU-Pro Medical subset (health and biology tracks, donate as `MMLU-Pro(M)')~\citep{wang2024mmlupro}. We also conducted tests on the expert-level benchmark MedXpertQA~\citep{zuo2025medxpertqa}, divided into two parts: Understanding (MedXpert-U) and Reasoning (MedXpert-R) , which are used to separately evaluate the improvement of our method on different types of tasks.

\paragraph{Models, Evaluation and Training.} We evaluated Med-REFL's ability to enhance reasoning capabilities across a diverse suite of six 7B-8B parameter baseline models. To better understand its impact, we grouped these models into three categories based on their primary architectural strengths and training focus. First,  the \textbf{Instruction-Tuned} models (Llama3.1-8B~\citep{grattafiori2024llama} and Qwen2.5-7B~\citep{qwen2025qwen25technicalreport}) are selected to assess our framework's effectiveness at instilling specialized skills from a generalist base. Second, the \textbf{Reason-Heavy} models (Huatuo-o1-8B and DeepSeek-Distill-8B~\citep{deepseekai2025deepseekr1incentivizingreasoningcapability}), are specifically optimized for complex logical deduction, allowing us to test Med-REFL's capacity to refine already strong reasoning faculties. Third, the \textbf{Knowledge-Heavy} models (MedReason-8B~\citep{medreason} and UltraMedical3.1-8B~\citep{zhang2024ultramedical}) are extensively fine-tuned on vast medical corpora, which are to evaluate the framework's impact on models rich in domain-specific knowledge. In addition to these SFT-trained models, the model AlphaMed~\citep{liu2025distillationpushinglimitsmedical}, trained entirely with GRPO~\citep{shao2024deepseekmath}, was also tested. To ensure stable and reliable results, all reported scores are the average of three independent runs. More evaluation and training details could be found in Appendix~\ref{subapdx:supplementary}.

\paragraph{Training Dataset Construction.} 
Our preference dataset was constructed from the MedQA-USMLE training set via a two-stage automated pipeline. A detailed description of the entire data construction process, including the creation of different preference pair types, is available in Appendix~\ref{subapdx:data_construction}.

\subsection{Main Results and Analysis}

The comprehensive results presented in Table~\ref{tab:comprehensive_results} validate the potent efficacy of our Med-REFL framework. Our methodology delivers consistent and significant performance improvements across both the in-distribution MedQA benchmark and a diverse suite of OoD challenges. Notably, Med-REFL substantially elevates the performance of already SoTA specialized models (Huatuo-o1-8B, Avg. +3.59\%), demonstrating its value not just as a training method, but as a powerful enhancement framework for even the strongest existing models.

\paragraph{Analysis across Model Architectures.} A deeper analysis reveals a compelling pattern in how Med-REFL interacts with different model types. While it successfully instills complex reasoning skills in general-purpose instruction-tuned models and helps knowledge-heavy models better apply their vast knowledge, the most pronounced average gains are observed in the \textbf{Reason-Heavy} category like Huatuo-o1. By observing the different reasoning pattern in model's output, we hypothesize this is because these models, through their original training, have already developed a rudimentary framework for reflection. Med-REFL, in this context, acts as a powerful catalyst; it \textbf{hones and refines this pre-existing, latent capability}, leading to outsized performance improvements. Besides, it also proves its compatibility with pure reinforcement learning by consistently boosting AlphaMed's performance (e.g., +2.52\% on GPQA-Med and +1.55\% on MedQA).

\paragraph{Performance across Benchmark Characteristics.} The framework's impact is particularly significant on datasets that demand deep, multi-step deductive reasoning, such as GPQA (Med+) and MMLU-Pro (Med+), which saw average gains of +3.89\% and +2.37\% respectively across all models. This demonstrates that Med-REFL directly enhances the core logical faculties of the models. Furthermore, it showcases consistent gains on the MedXpert benchmark on the most challenging dataset, MedXpertQA and the improvement for understanding questions is significantly greater than for reasoning questions. In contrast, while still positive, the gains on datasets that are more retrieval-oriented, like PubMedQA, are more modest. This is expected, as such tasks rely less on the iterative, complex self-correction process that our framework is designed to teach, further validating that Med-REFL's primary contribution is to the quality of the reasoning process itself.

\begin{figure*}
    \centering
    \includegraphics[width=1\linewidth]{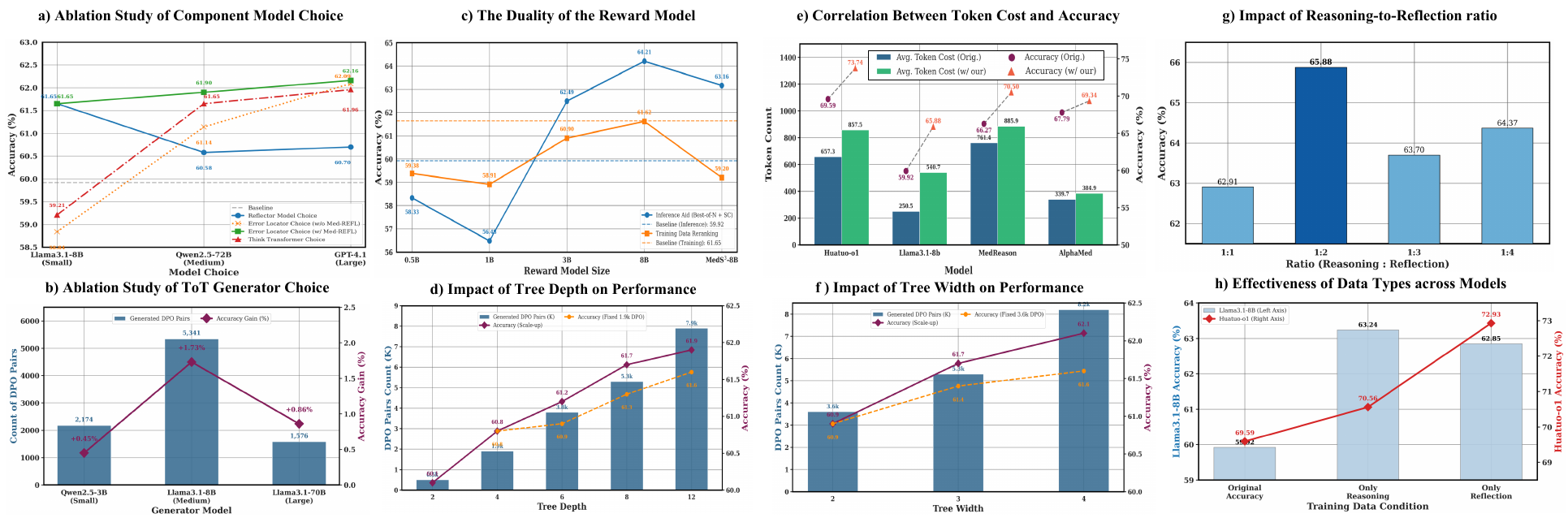}
    \caption{ Ablation and Analysis Experiments.
    a) The impact of different model choices for the Reflector, Error Locator, and Think Transformer($M$) on final accuracy.  
    b) The relationship between the generator size, the volume of valid DPO pairs constructed, and the resulting accuracy gain. 
    c) Effects of reward models with different sizes on inference (best-of-N + self-consistency) and training inference  
    d) \& f) Increasing tree depth(e) and width(f) results in DPO pairs numbers and performance.
    e) The relationship between average token cost per question and performance on MedQA(5 op).
    g) Impact of training data ratio.
    h) Independent effects of two types of data.
    }
    
    \label{fig:main_results_2}
    \vspace{-10pt}
\end{figure*}

\paragraph{Analysis of Reasoning Token Cost and Competitive Performance.}
We further analyzed the impact of Med-REFL on the nature of the reasoning process itself. As illustrated in Figure~\ref{fig:main_results_2} a), it reveals a strong positive correlation between the length of a model's reasoning trace and its final accuracy on MedQA. The application of Med-REFL consistently led to longer, more detailed reasoning, which is structurally linked to the performance gains. Furthermore, detailed results available in Appendix ~\ref{subapdx:fig_results} highlights the exceptional standing of our Med-REFL-8B model (Huatuo-o1 trained with Med-REFL). This proves that our framework provides a resource-efficient pathway to achieving top-tier medical reasoning capabilities without relying on massive model scale.

\section{Ablation Studies and Further Analysis}
\label{sec:ablations}
To investigate the contribution of each component to Med-REFL's performance, we used the MedQA-USMLE development set for training and conducted a series of ablation experiments on the test set for validation.
\begin{itemize}
    \item \textbf{Model selection for components:} For $M$, $R$, and $E$, we performed ablation experiments using the Small Model (Llama3.1-8b), Medium Model (Qwen2.5-72b), and Large Model (GPT4.1), respectively, with the results presented in Fig. ~\ref{fig:main_results_2}a). For the ToT data generator, we also conducted tests with open-source models of different sizes, as shown in Fig.~\ref{fig:main_results_2}b).
    \item \textbf{Impact of Tree Search hyperparameters:} We set the tree search depth to 2, 4, 6, 8, and 12, and the tree search width to 2, 3, and 4, respectively, to generate rollout data. Results are shown in Fig.~\ref{fig:main_results_2} d) and f). The dashed line represents the training performance when the number of pairs is fixed (randomly sampled from the total number). We observed an efficiency `sweet spot' at a width of 3 and a depth of 8. Beyond these thresholds, further expansion of the search space yields diminishing returns, with no significant improvement in the final accuracy.
    \item \textbf{Impact of training data composition:} To explore how the ratio of reasoning-augmented and reflection-augmented data affects model performance, we conducted relevant experiments with Llama3.1-8b, as presented in Fig.~\ref{fig:main_results_2} g). To investigate the independent effects of these two types of data, we trained Huatuo-o1 and Llama3.1-8b on each type separately, with the results shown in Fig. 3 h). It should be noted that the training dataset for these two experiments was aligned with that of the main experiment.
\end{itemize}

\subsection{\revise{Computational Cost and Efficiency Analysis}}
\label{subsec:cost_analysis}

\revise{Beyond performance gains, economic efficiency is also critical for medical LRM. Conventional approaches often incur prohibitive costs: distillation from closed-source models like Huatuo-o1 involves substantial API consumption (the researchers noted this as a considerable expense~\citep{huatuogpto1}), while paradigms like Med$^3$ require training separate, parameter-heavy reward models, consuming at least 112 GPU hours even without accounting for the undisclosed data generation time~\citep{Med333}. In contrast, Med-REFL offers a transparent and highly cost-effective solution, as detailed in Table~\ref{tab:cost_comparison}. Our entire pipeline, consisting of local data generation( from MedQA training set to Med-REFL DPO dataset,$\approx$83h) and standard DPO training ($\approx$12h) on a single A100, achieves a superior average gain of +3.59\% on the SoTA model (Huatuo-o1). This demonstrates that Med-REFL not only reduces resource dependency on external APIs and auxiliary reward models but also delivers more robust performance improvements.}
\begin{table}[htbp]
    \centering
    \vspace{-5px}
    \small
    \captionsetup{skip=4pt}
    \caption{\revise{Cost comparison between different reasoning enhancement paradigms. Estimated costs are based on standard A100 cloud rental rates (\$2 / GPU $\times$ hour) or public API pricing.}}
    \label{tab:cost_comparison}
    
    \resizebox{\linewidth}{!}{%
        \begin{tabular}{lllc}
        \toprule
        \textbf{Method} & \textbf{Resource Dependency} & \textbf{Est. Time / Cost} & \textbf{Average Gain} \\
        \midrule
        Huatuo-o1 & Distillation from GPT-4o (Paid API) & \$1,000 ($\approx$500GPU Hours) & +3.9 (on Llama3.1-8b) \\
        \midrule
        \multirow{2}{*}{Med$^3$} & Training and Inferencing & 112 GPU hours (Only training time, & \multirow{2}{*}{+1.98 (on Llama3.1-8b, Bo4)} \\
         & an additional Reward Model & data generation was not disclosed) & \\
        \midrule
        \textbf{Med-REFL} & \textbf{Local Deployed Data Generation} & \textbf{$\approx$83 GPU Hours} & \multirow{2}{*}{\textbf{+3.59 (on SoTA model)}} \\
        \textbf{(Ours)} & \textbf{Standard DPO} & \textbf{$\approx$12 GPU Hours} & \\
        \bottomrule
        \end{tabular}%
    }
    \vspace{-15px}
\end{table}

\subsection{Generalization Beyond Medical Multiple-choice Questions}
\label{subsec:gen_beyond}
To investigate the generalizability of our framework beyond constrained medical exams, we extend our evaluation to two distinct reasoning paradigms: the Knights and Knaves (K\&K)~\citep{xie24memorization} and the DDXPlus~\citep{fansi2022ddxplus}. While K\&K assesses logical deduction in a non-medical domain with exponentially scaling difficulty ($n$), DDXPlus focuses on diagnostic reasoning within electronic health records, testing the framework's adaptability to  realistic clinical scenario. Detailed experimental configurations for both tasks are provided in Appendix~\ref{apdx:gen_setup}.
As summarized in Table~\ref{tab:gen_beyond}, the results are compelling. For logic puzzles, the results trained solely on medical data (+Med-REFL, avg gain 2.98\% ) suggests that the self-correction capability is a transferable cognitive skill rather than mere domain memorization. Furthermore, specialized training (+Domain-REFL) yields substantial gains, particularly on unseen complex $n=4$ puzzles (+9.94\%). Similarly, results on the DDXPlus confirms that Med-REFL is a domain-agnostic and format-independent framework capable of unlocking complex reasoning across diverse scenarios.

\begin{table}[htbp]
    \centering
    \small
    \captionsetup{skip=4pt}
    \caption{Generalization performance across diverse reasoning domains and formats. \textit{+Med-REFL} denotes zero-shot transfer from medical MCQ data, while \textit{+Domain-REFL} indicates fine-tuning on task-specific generated reflection data.}
    \label{tab:gen_beyond}
    \resizebox{\linewidth}{!}{
        \begin{tabular}{llccc}
        \toprule
        \textbf{Domain} & \textbf{Setting} & \textbf{Llama3.1-8B} & \textbf{+Med-REFL} & \textbf{+Domain-REFL} \\
        \midrule
        \multirow{4}{*}{Logic (K\&K)} 
        & $n=2$ 
            & \res{38.77}{0.60} 
            & \res{40.82}{0.26} 
            & \res{\textbf{49.93}}{0.48} \\
        & $n=3$ 
            & \res{22.88}{0.22} 
            & \res{29.23}{0.30} 
            & \res{\textbf{40.11}}{0.57} \\
        & $n=4$ 
            & \res{13.40}{0.17} 
            & \res{13.96}{0.98} 
            & \res{\textbf{23.34}}{0.83} \\
        \cmidrule{2-5}
        & \textbf{Avg (Logic)} 
            & {25.02} 
            & {28.00} 
            & {\textbf{37.79}} \\
        \midrule
        EHR Diagnosis & DDXPlus 
            & \res{27.60}{0.67} 
            & \res{29.30}{0.15} 
            & \res{\textbf{32.10}}{1.26} \\
        \bottomrule
        \end{tabular}
    }

\end{table}

\subsection{\revise{Comprehensive Ablation and Robustness Analysis}}
\revise{Due to space constraints, we present extensive additional validations in the appendix. Regarding data construction, we verify the superiority of ToT-based exploration over random rollouts (Appendix~\ref{apdx:Rollout_Data_Generation}) and demonstrate the necessity of contrastive learning by comparing DPO against SFT (Appendix~\ref{secc:prompt_ablation}). Furthermore, we provide additional hyperparameter(temperature, random seed and $\lambda$) sensitivity analysis in Appendix~\ref{appendix:sensitivity}.}

\subsection{Qualitative Analysis}
\begin{figure}
    \centering
    \includegraphics[width=1\linewidth]{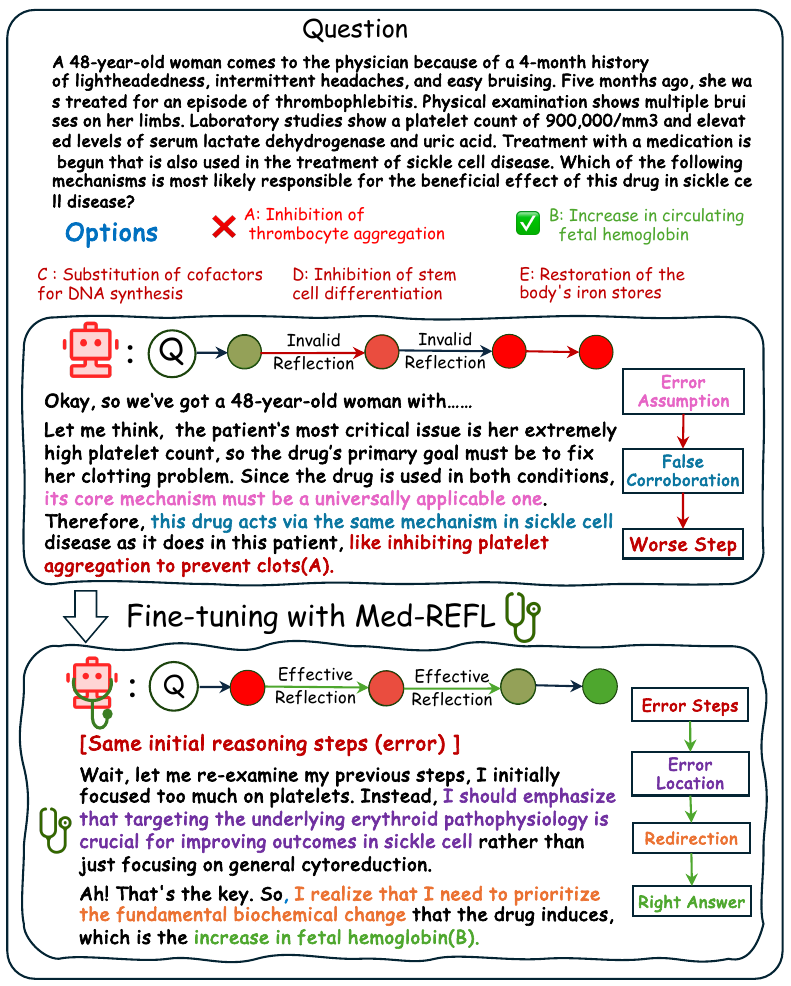}
    \caption{Case study: how Med-REFL teaches huatuo-o1 to reflect.}
    \label{fig:casestudy}
    \vspace{-10px}
\end{figure}

To complement our quantitative results, Figure~\ref{fig:casestudy} offers a qualitative snapshot of Med-REFL's impact on the model's internal reasoning process. More detailed case studies and analysis are provided in Appendix~\ref{apdx:case_study} for a deeper intuitive understanding.

\section{\revise{Discussion}}
\label{sec:discussion}

\subsection{Probing Meta-Cognitive Reflection and Error Recovery}

We conduct a multi-dimensional diagnostic analysis to discern whether Med-REFL instills a genuine self-reflection capability rather than merely encouraging longer reasoning traces and exhibiting ``fake reflection'' (confirmatory bias), where models predominantly reaffirm initial errors rather than genuinely correcting them\citep{kang2025first}.
We conduct controlled stress tests on the MedQA-USMLE test set. First, we construct two balanced diagnostic subsets, each containing 500 samples: (1) \textbf{$\mathcal{D}_{\text{intrinsic}}$}, comprising low-value ($v_{\text{sol}}-v_{\text{rem}}$) reasoning steps identified during Llama3.1-8b's ToT rollouts, representing internal logic failures; and (2) \textbf{$\mathcal{D}_{\text{adversarial}}$}, where error initial steps are synthesized by Gemini-3-flash to mislead the model. In both settings, the models are forced to continue the reasoning process from a flawed starting point. 

Second, to verify whether Med-REFL functionally breaks this confirmation loop or merely induces stochastic instability, we conduct another ablation study using a specialized \textit{MedQA-Rethink} dataset. We collect trajectories to rollout from the MedQA-USMLE test set where the initial reasoning led to an incorrect answer. The model is then forced to revise its judgment by appending the prompt: ``Wait! I need to carefully rethink my earlier thought.'' We compare the base models only trained with reasoning enhancement (+RE) and the full Med-REFL dataset(+Refl). 
We define two dynamic metrics to evaluate the quality of this internal revision: (1) Reflection Divergence Rate, which measures how frequently the model alters its final answer after rethinking ($A_{think} \neq A_{reflect}$); and (2) Correction Success Rate, which calculates the conditional probability that an answer change successfully rectifies the initial error(($A_{reflect}=Correct | Changed$)).

As shown in Table~\ref{tab:combined_ablation}, standard reasoning enhancement (+RE) often leads to performance degradation or reinforces confirmatory bias, suggesting that pure reasoning data may make models more vulnerable to initial errors. In contrast, the full Med-REFL framework (+Refl) consistently achieves the highest recovery accuracy and nearly doubles the correction Success Rate for Huatuo-o1. These results confirm that the observed self-correction is a learned, transferable meta-cognitive skill rather than a simple byproduct of increased token length or domain memorization.

\begin{table}[htbp]
\centering
\caption{Comprehensive ablation study on error recovery and reflection effectiveness. We compare the base model (Original), reasoning enhancement (+RE), and reflection enhancement (+Refl) across diagnostic subsets and the Rethink benchmark.}
\label{tab:combined_ablation}
\resizebox{\columnwidth}{!}{
\begin{tabular}{@{}lcccccc@{}}
\toprule
\textbf{Model} & \multicolumn{3}{c}{\textbf{Llama 3.1-8B}} & \multicolumn{3}{c}{\textbf{Huatuo-o1}} \\ 
\cmidrule(lr){2-4} \cmidrule(lr){5-7} 
\textbf{Setting / Metric} & Original & + RE & + Refl (Ours) & Original & + RE & + Refl (Ours) \\ \midrule
\textit{Error Recovery (Acc \%)} & & & & & & \\
$\mathcal{D}_{int}$ (Intrinsic) 
    & \res{27.53}{1.22} & \res{25.47}{0.23} & \res{\textbf{33.43}}{0.61} 
    & \res{46.17}{2.48} & \res{44.97}{0.52} & \res{\textbf{56.93}}{0.37} \\
$\mathcal{D}_{adv}$ (Adversarial) 
    & \res{14.60}{0.96} & \res{11.57}{1.45} & \res{\textbf{17.87}}{0.42} 
    & \res{25.87}{0.12} & \res{24.64}{0.40} & \res{\textbf{30.50}}{0.87} \\ \midrule
\textit{MedQA-Rethink (\%)} & & & & & & \\
Divergence Rate 
    & \res{5.38}{2.68} & \res{5.47}{1.37} & \res{\textbf{10.13}}{3.04} 
    & \res{9.45}{2.57} & \res{8.91}{3.81} & \res{\textbf{14.20}}{1.12} \\
Success Rate 
    & \res{7.87}{0.75} & \res{6.92}{1.02} & \res{\textbf{14.54}}{0.99} 
    & \res{12.68}{1.26} & \res{11.39}{2.17} & \res{\textbf{21.49}}{1.28} \\ \bottomrule\\ 
\end{tabular}
}

\end{table}

\subsection{Large Teacher Model is not always better}
\label{sec:discussion_model_choice}

To disentangle the impact of model capacity within the Med-REFL framework, we conducted a comprehensive component-level ablation study using Llama3.1-8B as the base student model on the MedQA-USMLE development set. Contrary to the prevailing assumption that larger teacher models invariably yield superior distillation signals, our experiments, as illustrated in Figure~\ref{fig:main_results_2} a) \& b), reveal a nuanced landscape where the optimal model choice depends heavily on the specific functional role within the pipeline.

We first examine the `Reflector', which generates the reflection trace. As shown in the blue line of Figure~\ref{fig:main_results_2} a), the student model itself (Llama3.1-8B) unexpectedly outperforms stronger reflector models. This suggests that reflection derived from the model's own latent space are more readily internalized than the larger models' reasoning with significant distribution differences. Similar phenomena have been independently reported in recent studies~\citep{li-etal-2025-small-models, cai-etal-2025-enhancing, lai2024stepdpo}.

In terms of the `Error Locator', the impact of model scale is effectively neutralized by our structural valuation mechanism. Figure~\ref{fig:main_results_2} a) demonstrates that without Med-REFL, the intuition of the 8B model is insufficient to locate errors, significantly behind GPT-4.1. However, when augmented with Med-REFL's deterministic value estimation (Equation ~\ref{eq:serro}), the 8B model achieves parity with GPT-4.1. This indicates that our algorithm successfully bridges the capability gap, enabling smaller models to perform reliable diagnostics without relying on expensive external supervision.

Conversely, the `Think Transformer (Thinker, $M$)', responsible for verbalizing the structured path into a natural language stream, exhibits a strict positive correlation with model scale. The red dashed line in Figure~\ref{fig:main_results_2} a) shows a sharp performance drop when using the 8B model (59.21\%) compared to GPT-4.1 (61.96\%). We attribute this to the high complexity of the rewriting task; larger models possess better instruction-following and linguistic capabilities to preserve the logical fidelity of the reasoning path during verbalization, whereas smaller models prone to lose information.

Finally, regarding the `ToT Generator', Figure~\ref{fig:main_results_2} d) reveals a sweet pot phenomenon. The medium-sized Llama3.1-8B achieves higher accuracy gains (+1.73\%) than both smaller and larger choices. The 3B model is too weak to generate plausible correct paths, while the 70B model is too robust, rarely producing the errors necessary to construct the `rejected' samples for contrastive learning. 

\subsection{Reward Model is useful in inferencing but not in training}
\label{sec:discussion_reward_model}

To further delineate the role of value estimation in our framework, we investigated the potential of training a separate reward model to generalize the step-wise solution values derived from our ToT exploration. We hypothesized that a parameterized RM might offer a more robust signal for both inference-time and training data re-ranking. We trained a series of reward models ranging from 0.5B to 8B parameters on the collected trajectory data and evaluated their impact on two distinct stages: inference (via Best-of-N reranking) and training (as a data selector for DPO).
The results, visualized in Figure~\ref{fig:main_results_2} c), present a striking dichotomy. During the inference stage, the RM proves to be a powerful tool. 
However, a counter-intuitive phenomenon emerges when applying these RMs to the training stage. Contrary to this expectation, none of the results exceeded the baseline.
We conclude that while reward Models are beneficial for inference, our intrinsic structural evaluation already provides the optimal, noise-free signal for self-correction learning, making auxiliary RMs unnecessary for training.

\section{Conclusion}
In this work, we introduced Med-REFL, a novel framework designed to tackle the expensive cost of verifying intermediate medical reasoning. It leverages a deterministic, structure-based method to automatically construct targeted preference data, effectively enhancing a model's self-reflection capabilities. Our experiments demonstrate that Med-REFL achieves SoTA performance on the MedQA-USMLE benchmark and exhibits robust generalization across diverse medical datasets, with ablation studies confirming that it instills a genuine internalized self-correction capability. Furthermore, we believe that the core principles of Med-REFL are applicable to other domains facing similar verification bottlenecks, such as legal reasoning, and exploring this transferability will be a key direction for our future work. Our findings show that explicitly training for a high-quality internalized reflection capability is a crucial and effective pathway toward developing more reliable and trustworthy LRMs for high-stakes medical applications.

\section*{Impact Statement}

This paper introduces Med-REFL, a framework designed to enhance the reliability of LRMs in medical reasoning. By addressing the `verification bottleneck' through deterministic structural assessment and preference learning, our work has several potential broader impacts:

\textbf{Societal Benefits:} The most immediate benefit is the development of more trustworthy AI assistants for clinical decision support. By internalizing a self-correction mechanism, Med-REFL enhances the model's ability to recover from initial errors and reduces errors in resource-constrained settings where access to specialist expertise is limited. Furthermore, the framework provides a cost-effective path to building specialized models without relying on expensive human-expert annotations, democratizing the development of high-performance medical AI.

\textbf{Ethical Considerations and Risks:} Despite the improvements in reasoning robustness, two primary risks remain. First is automation bias: as models become more adept at generating `reflective' and detailed reasoning traces, users may be lulled into a false sense of security and over-rely on the model's outputs. Second, while Med-REFL mitigates `fake reflection,' it does not eliminate the possibility of sophisticated errors that align with the model's internal logic but remain clinically incorrect. If used as a primary diagnostic tool without human oversight, such errors could lead to delayed or incorrect treatments.

\textbf{Mitigation and Initiatives:} We emphasize that Med-REFL is intended to function as a collaborative tool for medical professionals, rather than a replacement for clinical judgment. To mitigate potential risks, we encourage the community to: (1) conduct further research into human-AI interaction to understand how reflective reasoning traces affect clinician behavior; (2) implement `human-in-the-loop' safeguards when deploying these models in real-world clinical workflows; and (3) continue investigating the `reasoning boundary' to identify specific medical sub-domains where the model's self-correction capabilities might still be insufficient.


\bibliography{example_paper}
\bibliographystyle{icml2026}

\newpage

\appendix

\clearpage

\section{Related Works}
\label{related_works}

\subsection{Outcome-Oriented Reasoning Enhancement}
The most established paradigm for enhancing large language model (LLM) reasoning involves supervision based on final outcomes. This is primarily achieved through two training avenues: SFT and RL. In the medical domain, SFT is widely used to train models on high-quality chain-of-thought (CoT) data distilled from powerful teacher models, as seen in works like Huatuo-o1 and ReasonMed~\citep{reasonMed}. Concurrently, outcome-based RL method such as GRPO~\citep{shao2024deepseekmath} and DAPO~\citep{yu2025dapoopensourcellmreinforcement} refines model policies by rewarding the final correct answer, a technique successfully transferred to medicine by Med-RLVR~\citep{zhang2025med} and AlphaMed~\citep{liu2025distillationpushinglimitsmedical}, and foundational to general reasoners like DeepSeek-R1~\citep{deepseekai2025deepseekr1incentivizingreasoningcapability}. While effective at improving task-specific performance, these outcome-oriented methods share a critical vulnerability: the learning signal is sparse and lacks process-level granularity~\citep{zhang2023chainofthoughtreasoningpolicyimprovement}. This makes them susceptible to `reward hacking', where flawed logic that coincidentally produces a correct answer is positively reinforced~\citep{lightman2023let}. Furthermore, the reliance on powerful teacher models for high-quality data distillation is often prohibitively expensive~\citep{cost}. This paradigm, therefore, struggles to guarantee the learning of robust and sound reasoning processes.

\subsection{Process-Oriented Reasoning Enhancement}
To address the lack of intermediate supervision, PRMs have been proposed to evaluate each step in a reasoning chain. This approach has been adapted for medical reasoning by training PRMs on stepwise quality labels, as demonstrated in Med-PRM~\citep{medprm} and MedS³~\citep{Med333}. These labels are often generated via complex methods, such as statistical approximation from massive random rollouts~\citep{prmrandomrollout,Med333} or judgments from a complex retrieval-augmented generation (RAG) system~\citep{medprm}. In some cases, as with MedReason~\citep{medreason}, the reasoning process is constrained by a knowledge graph to ensure factual grounding. However, this granularity comes at a significant cost. The PRM approach typically requires training a separate, powerful reward model, introducing considerable computational overhead and architectural complexity~\citep{yuan2024self}. More fundamentally, these frameworks are often optimized to identify the most correct reasoning path. This objective is insufficient for teaching a model how to reflect and recover from its own diverse errors~\citep{huang2023large}, hindering its ability to develop a robust, generalizable self-correction capability~\citep{lan2024training}.

\subsection{Reasoning Path Exploration and Evaluation}
A third line of work leverages tree-search algorithms like monte carlo tree search (MCTS)~\citep{kocsis2006bandit} and ToT to explore a vast space of reasoning possibilities as training data. This approach is exemplified by methods such as rStar-math~\citep{guan2025rstar}, ASTRO~\citep{kim2025astro}, and the O1-Journey~\citep{o1-part1,o1-part2} series. These methods are powerful for generating diverse reasoning data, including paths that contain trial-and-error, which is crucial for learning complex problem-solving. The primary limitation of many such approaches lies in the evaluation of the generated steps. Node values are often approximated via noisy statistical estimations from random rollouts~\citep{hao2023reasoning}, and the exploration can lead to `ineffective reflections', jumps from one flawed reasoning state to another, without a guaranteed improvement in reasoning quality~\citep{huang2023large}. While these methods explore extensively, they often lack a robust mechanism to assess the value of a corrective action itself based on global information~\citep{hao2023reasoning}. In contrast, Med-REFL's deterministic, structure-based assessment is specifically designed to quantify the value of these corrective actions, ensuring that the learned reflection is consistently effective.

\subsection{\revise{Self-Reflection and Confirmatory Bias in LRMs}}
\revise{Recent studies highlight a critical limitation in the reasoning capabilities of LRMs: the inability to genuinely self-correct due to deep-seated confirmatory bias. \cite{kang2025first} empirically demonstrate that model-generated reflections are predominantly confirmatory (`fake reflection'), with the correctness of the final output largely dictated by the initial reasoning attempt rather than subsequent reflection. This phenomenon is underpinned by `reasoning inertia', where models develop an internal bias immediately upon processing the input, leading to `overthinking' and redundant steps that reinforce the initial error rather than correcting it \cite{dang2025internal}. Consequently, without explicit supervision on \textit{how} to correct, the reflection process frequently devolves into a `post-hoc rationalization' of the initial failure \cite{zhang2025critique}. Unlike previous approaches that rely on stochastic sampling or weak critique, Med-REFL explicitly constructs preference pairs to distinguish `effective correction' from `invalid reflection', directly supervising the model to break this confirmatory loop.}

\section{Method Reproducibility Statement}
\label{apdx:reprodu}
The code, data, and weights have been open-sourced and are available for reproduction. \\ Readers can visit the anonymous repository at: \\ https://anonymous.4open.science/r/Med-REFL-B5F8/ \\ All the information has been anonymized, and the complete Med-REFL training dataset can be found in the repository's `Data' folder.

\section{Supplementary Methods and Experiments}

\subsection{Tree-of-Thought Path Searching for Medical Reasoning}
\label{subapdx:tot}
This section details the tree-of-thought (ToT) based path searching mechanism employed to generate diverse reasoning trajectories for complex medical problems. This process of structured exploration forms the foundational stage for subsequent fine-grained evaluation and preference data construction within our Med-REFL framework. The ToT approach decomposes a given medical problem into a sequence of discrete reasoning steps or intermediate thoughts, which collectively form a tree-like structure. In this structure, each node represents an intermediate reasoning state, and a path from the root to any leaf node constitutes a complete reasoning trajectory towards a potential solution.

Formally, the generation of such a reasoning trajectory is an iterative process. Let $Q$ represent the initial medical problem description. The LLM, denoted as $\pi$, serves as the core reasoning engine. The process initiates by generating an initial reasoning step (or thought), $s_0$, based on $Q$:
\begin{equation}
\label{eq:appendix_tot_s0}
s_{0} \leftarrow \text{ProposeInitialStep}(Q, \pi)
\end{equation}
Subsequent reasoning steps are generated iteratively, conditioned on the original problem $Q$ and the history of preceding steps. If $H_t = (s_0, s_1, \dots, s_{t-1})$ denotes the sequence of thoughts generated up to step $t-1$, the next thought $s_t$ is proposed as:
\begin{equation}
\label{eq:appendix_tot_st}
s_{t} \leftarrow \text{ProposeNextStep}(Q, H_t, \pi)
\end{equation}
This formulation ensures the LLM $\pi$ considers both the original problem and the accumulated reasoning history to maintain context and coherence. The process continues until a candidate solution is formed or a predefined maximum length $k_{max}$ is met. A complete reasoning trajectory leading to a candidate answer $a_Q$ is thus represented by the sequence $S_{final} = (s_0, s_1, \dots, s_{F})$, where $F < k_{max}$.

The prompts used for constructing ToT paths can be found in supplementary materials or code.

\subsection{Detailed Description of Data Construction}
\label{subapdx:data_construction}

Our preference dataset was constructed exclusively from the official MedQA-USMLE training set. The entire process was driven by a two-stage automated pipeline, which is detailed below. The official test split of the benchmark was reserved strictly for final performance evaluation.

\paragraph{Stage 1: Seed Trajectory Generation.}
The initial step focused on generating a diverse set of reasoning paths.
\begin{itemize}
    \item We utilized Llama3.1-8B with a ToT exploration strategy to process the MedQA training set. This generated an initial pool of 45,000 reasoning trajectories.
    \item From this pool, we filtered down to 28,000 valid trajectories that contained a mix of both correct and incorrect reasoning paths.
    \item Each of these 28,000 trajectories was assigned a score based on its solution value, creating a dataset we denote as $D_{\text{SFT}}$. This dataset was subsequently used to train the scoring model introduced in Section~\ref{sec:discussion_model_choice}.
\end{itemize}

\paragraph{Stage 2: High-Quality Preference Pair Construction.}
In the second stage, the 45,000 initial trajectories were first reconstructed into 10,000 distinct reasoning trees. These trees then served as the foundation for constructing our final DPO preference pairs. The dataset was partitioned by question, with approximately one-third of the questions allocated for `Reasoning Enhancement' data and the remaining two-thirds for `Reflection Learning' data. 
\begin{itemize}
    \item \textbf{Reasoning Enhancement Pairs (12k):} These pairs were designed to teach the model general reasoning proficiency by contrasting high-quality correct reasoning paths against plausible but flawed ones.
    \item \textbf{Reflection Learning Pairs (21k):} These pairs specifically train the model's self-correction capabilities. They were generated by identifying an error in a reasoning path and contrasting an \textit{ineffective reflection} (which fails to correct the error) with an \textit{effective reflection} (which successfully identifies and corrects the error). The generation logic for these pairs was as follows:

If the \textit{Error Locator} identified an incorrect step at a non-leaf node of the reasoning tree, a \textbf{mid-reasoning} reflection pair was generated.

If the identified error was at a leaf node, we generated both a \textbf{mid-reasoning} pair (correcting the flawed leaf node to a correct one) and a \textbf{post-reasoning} pair (reflecting on the entire flawed trajectory from the error onward).
\end{itemize}
The Error Locator and Reflector, were executed by the Llama3.1-8b model via the vLLM~\citep{kwon2023efficient} engine. Qwen2.5-72B-Int4 were used as thinker ($M$). 

\subsection{Supplementary Experiments Settings}
~\label{subapdx:supplementary}
\paragraph{Key hyperparameters}
 Key hyperparameters for our value functions were set as follows: for the action value ($v_{act}$), the weights were $\lambda_1=0.4, \lambda_2=0.2, \lambda_3=0.4$; 

\paragraph{Evaluation and Training Details}
For inference, we used a maximum generation length of 8192 tokens and randomized key sampling parameters, with temperature and $\text(top_p)$ each drawn from the range [0.2, 1.0], to assess robust performance across various generation conditions. 
For efficient fine-tuning, all baseline models were adapted using LoRA~\citep{hu2022lora}  with a rank $r=64$ and alpha $\alpha=16$. Models were trained for 1 epoch with a learning rate of $1 \times 10^{-5}$ and a global batch size of 128. We utilized the LlamaFactory~\citep{zheng-etal-2024-llamafactory} framework for training, which was optimized using DeepSpeed-ZeRO~\citep{zero2020samyam} Stage 3. All fine-tuning experiments were successfully conducted on a standard setup of 2 NVIDIA A100 (80GB) GPUs.

\subsection{Experimental Settings for Open-ended Generalization Tasks}
\label{apdx:gen_setup}

\paragraph{Logic Puzzles (K\&K)} 
For the K\&K benchmark, we utilized the standard test sets for $n \in \{2, 3, 4\}$ and constructed a specialized `Puzzle-REFL' dataset by seamlessly transferring our ToT framework: using training set ($n=2,3$) and adapting the ToT prompts to logic solvers.

\paragraph{EHR Diagnosis (DDXPlus)} 
Unlike standard medical exam datasets, DDXPlus\citep{fansi2022ddxplus} is a large-scale EHR dataset featuring millions of synthetic patient encounters designed for machine learning evaluation. Due to its scale, exhaustive LLM evaluation is impossible and reveals significant variance across different research~\citep{Med333, medprm}. Following the methodology of Med$^3$~\citep{Med333}, we use a subset of 2,000 cases that encompasses all 49 potential diagnoses. We utilized 1,000 samples for reflection-based fine-tuning and the remaining 1,000 samples for the final performance evaluation. This setup ensures that we evaluate the model's ability to handle complex, high-resolution clinical data beyond simple multiple-choice formats.

\subsection{Detail Results about Token Cost and Med-REFL-8b's Competitive Performance}
~\label{subapdx:fig_results}
This section provides the detailed numerical data and corresponding analysis for the trends and competitive performance.

\paragraph{Analysis of Reasoning Depth}
\begin{table}[h]
  \centering
  \sisetup{table-format=4.2} 
  \caption{Comparison of model performance on MedQA, showing average token cost per question and accuracy before and after applying our method.}
  \label{tab:token_cost_appendix}
  \resizebox{\linewidth}{!}{%
  \begin{tabular}{l S S S[table-format=2.2] S[table-format=2.2]}
    \toprule
    & \multicolumn{2}{c}{\textbf{Avg. Token Cost}} & \multicolumn{2}{c}{\textbf{Accuracy (\%)}} \\
    \cmidrule(lr){2-3} \cmidrule(lr){4-5}
    \textbf{Model} & {\textbf{Original}} & {\textbf{w/ our}} & {\textbf{Original}} & {\textbf{w/ our}} \\
    \midrule
    Huatuo-o1             & 657.30  & 857.52  & 69.59 & 73.74 \\
    Llama3.1-8B           & 250.46  & 540.73  & 59.92 & 65.88 \\
    MedReason             & 761.35  & 885.89  & 66.27 & 70.50 \\
    AlphaMed              & 339.74  & 384.95  & 67.79 & 69.34 \\
    Qwen2.5-7B            & 252.14  & 239.55  & 57.11 & 60.05 \\
    Deepseek-Distill-8B   & 1043.82 & 1530.12 & 48.85 & 55.07 \\
    Ultra-Medical         & 517.79  & 618.47  & 71.34 & 73.43 \\
    \bottomrule
  \end{tabular}
  }
\end{table}

As discussed in the main text, Table~\ref{tab:token_cost_appendix} demonstrates a clear relationship between reasoning depth and performance on MedQA. Across most evaluated models, applying Med-REFL leads to simultaneous increases in average token cost and inference accuracy, indicating that performance gains are structurally driven by more elaborate reasoning. Notably, Qwen2.5-7B constitutes an exception, showing improved accuracy despite a slight reduction in token cost, suggesting enhanced reasoning efficiency. Overall, these results confirm that Med-REFL improves model performance primarily by fostering deeper and more effective reasoning.

\paragraph{Competitive Performance Analysis}

\begin{table*}[h]
  \centering
  \caption{
    Performance comparison of Med-REFL-8B against other models. 
    Our model (\textbf{Med-REFL-8B}) is placed at the bottom of the 8B and 10-20B categories for direct comparison, separated by a dashed line. 
    The best score within each category (8B models and 10-20B models) is bolded.
  }
    \label{tab:main_results_appendix}
  
  \resizebox{\textwidth}{!}{%
  \begin{tabular}{@{}l S[table-format=2.2] S[table-format=2.2] S[table-format=2.2] S[table-format=2.2] S[table-format=2.2] S[table-format=2.2] S[table-format=2.2]@{}}
    \toprule
    \textbf{Model} & {\textbf{MedQA}} & {\textbf{MedMCQA}} & {\textbf{PubMedQA}} & {\textbf{GPQA}} & {\textbf{MMLU-Pro}} & {\textbf{MedXpert-U}} & {\textbf{MedXpert-R}} \\
    & {\textbf{(5op)}} & & & {\textbf{(M)}} & {\textbf{(M)}} & & \\
    \midrule
    \multicolumn{8}{@{}l}{\textit{\textbf{Reference Model ($>$20B)}}} \\
    Qwen-3-30B-a3B      & 81.74 & 69.83 & 74.20 & 52.31 & 77.95 & 21.73 & 19.88 \\
    \midrule
    \multicolumn{8}{@{}l}{\textit{\textbf{10B-20B Models}}} \\
    Qwen-3-14B          & \textbf{79.03} & \textbf{68.26} & 73.90 & 54.62 & \textbf{72.81} & 19.86 & \textbf{20.20} \\
    Microsoft-Phi-3-14B           & 69.60 & 62.92 & 69.10 & 52.05 & 69.77 & 17.66 & 14.08 \\
    Google-Gemma-3-12B & 65.67 & 58.00 & 60.60 & 51.03 & 68.14 & 16.81 & 14.24 \\
    Qwen-2.5-14B        & 65.12 & 61.55 & 73.43 & 53.50 & 59.64 & 13.92 & 12.95 \\
    Mistral-Nemo-14B    & 47.45 & 48.72 & 63.35 & 39.74 & 52.08 & 13.75 & 12.01 \\
    \textbf{Med-REFL-8B \textit{(for comp.)}} & {73.74} & {65.00} & \textbf{{81.60}} & \textbf{{57.01}} & {65.08} & \textbf{{20.09}} & {18.06} \\ 
    \midrule
    \multicolumn{8}{@{}l}{\textit{\textbf{General-purpose 8B Models}}} \\
    Qwen-3-8B           & 73.37 & 63.97 & 72.10 & 55.04 & \textit{75.83} & 19.19 & 16.28 \\
    Llama-3.1-8B        & 59.92 & 57.61 & 76.26 & 45.16 & 57.56 & 16.11 & 14.14 \\
    Qwen-2.5-8B         & 57.11 & 54.52 & 74.12 & 43.34 & 61.36 & 10.79 & 12.45 \\
    Mistral-8B          & 48.00 & 49.13 & 64.65 & 33.33 & 53.16 & 14.52 & 12.17 \\
    \midrule
    \multicolumn{8}{@{}l}{\textit{\textbf{Specially-trained 8B Models}}} \\
    UltraMedical-3.1-8B & 71.34 & 63.30 & 78.60 & 45.76 & 63.06 & 15.68 & 15.87 \\
    Huatuo-1-8B         & 69.59 & 62.13 & 79.02 & 50.67 & 61.87 & 15.31 & 16.85 \\
    AlphaMed-8B         & 67.79 & 61.22 & 78.03 & 53.40 & 63.29 & 20.24 & 21.91 \\
    MedReason-8B        & 66.27 & 58.98 & 77.17 & 45.64 & 59.14 & \textbf{22.55} & \textbf{22.31} \\
    Deepseek-Distill-8B & 48.85 & 49.51 & 69.30 & \textbf{62.43} & 53.83 & 13.07 & 11.98 \\
    \textbf{Med-REFL-8B} & \textbf{73.74} & \textbf{65.00} & \textbf{81.60} & 57.01 & \textbf{65.08} & 20.09 & 18.06 \\
    \bottomrule
  \end{tabular}%
  } 
\end{table*}
Table \ref{tab:main_results_appendix} collectively show that \textbf{Med-REFL-8b} demonstrates outstanding competitiveness both within its parameter class and against substantially larger models. Within the 8B regime, it achieves state-of-the-art performance on several challenging medical benchmarks, including MedQA, MedMCQA, PubMedQA, and MMLU‑Pro, confirming the effectiveness of the Med‑REFL framework in substantially enhancing a strong base model (Huatuo‑o1). More notably, \textbf{Med-REFL-8b} matches or surpasses many 10–20B models on multiple tasks, highlighting Med‑REFL as a resource‑efficient approach to attaining top‑tier medical reasoning capabilities without reliance on large‑scale parameter expansion.

\section{Supplementary Analyses and Ablation Studies}

\subsection{Analysis of ToT-based vs. Random Rollout Data Generation}
\label{apdx:Rollout_Data_Generation}
To validate Med-REFL's ToT-based data construction—a strategy designed to generate higher-quality, nuanced training signals than conventional methods—we compared its DPO data against pairs derived from a traditional random rollout ($RO$) approach. For the $RO$ baseline, multiple reasoning paths per question were sampled and scored using our 3B scoring model to select preference pairs ($RO_{Llama}$: 14k, $RO_{Huatuo}$: 10k). As shown in Table~\ref{tab:ablation_ro}, models trained with Med-REFL DPO data significantly outperform those trained with $RO$ data across both Huatuo-o1 (73.72\% vs. 69.97\%) and Llama3.1-8B (65.74\% vs. 62.21\%). This outcome highlights the superiority of Med-REFL's structured, ToT-based trajectory sampling and fine-grained evaluation. Such a principled approach is crucial for generating more effective preference pairs for complex reasoning, directly addressing the motivation to overcome limitations of less nuanced or purely result-oriented data generation strategies.

\begin{table}[htbp]
\centering
\caption{Ablation study on MedQA-USMLE comparing Med-REFL (our ToT-based DPO data) against DPO data from Random Rollouts (RO). Results highlight the performance advantage of Med-REFL's structured trajectory sampling and fine-grained evaluation over the RO approach.}
\label{tab:ablation_ro}
\resizebox{\linewidth}{!}{%
\begin{tabular}{lcc}
\toprule
Training Data & Huatuo-o1 & Llama3.1-8B \\
\midrule
Base Model & 69.59 & 59.92 \\
Random Rollout (RO) & 69.97 & 62.21 \\
\textbf{Med-REFL (Ours)} & \textbf{73.72} & \textbf{65.74} \\
\bottomrule
\end{tabular}
}
\end{table}

\subsection{Ablation on the Role of Negative Samples via Fine-tuning Strategy: DPO vs. SFT}
\label{secc:prompt_ablation}

We investigate which data components and learning mechanisms contribute most significantly to the performance gains. Our central hypothesis is that the key to enhancing reasoning lies not in merely imitating correct solutions, but in learning to distinguish correct reasoning from plausible but flawed alternatives. To test this, we conducted a crucial experiment comparing the effect of SFT against our full DPO approach. As shown in Table~\ref{tab:ablation_sft_dpo}, fine-tuning the model via SFT on only the positive examples (`chosen' paths) yielded a slight improvement. In stark contrast, the full Med-REFL approach, which uses DPO to learn from the preference pairs of `chosen' vs. `rejected' paths, delivered the full, substantial performance improvement. This result offers compelling evidence for our core thesis. It demonstrates that the dnegative samples identified by our framework are not noise, but a vital learning signal. The significant performance uplift is unlocked only through contrastive learning via DPO, which explicitly teaches the model to recognize and move away from erroneous reasoning patterns. Simply memorizing correct paths is insufficient; the model must learn what not to do. 
\begin{table}[h]
    \centering
    \small
    \caption{Ablation study on the fine-tuning method. ACC for accuracy. }
    \label{tab:ablation_sft_dpo}

    \begin{tabular}{lc}
    \toprule
    \makecell{Training Strategy \\ on Llama3.1-8B} & ACC (\%) \\
    \midrule
    Original (Base Model) & 59.92 \\
    + SFT (on all `chosen' paths) & 60.32 \\
    \textbf{+ DPO (Full Med-REFL)} & \textbf{65.74} \\
    \bottomrule
    \end{tabular}%

\end{table}

\subsection{Comparison with Alternative Strategies.}
\label{apdx:setup_comparison_strategies}

A central claim of our work is that Med-REFL's deterministic, structure-based evaluation provides a higher-quality learning signal than other paradigms. To verify this, we conducted a critical experiment comparing Med-REFL against two strong alternatives: a \textbf{statistical sampling} approach using monte carlo tree search (MCTS) style rollouts, and a powerful \textbf{LLM-as-a-judge} using GPT-4.1. To ensure a fair comparison, all methods operated on the exact same set of candidate reasoning steps generated during our initial ToT exploration; the only variable was the mechanism act as `action value' used to select the `chosen' vs. `rejected' paths for DPO. The results, presented in Table~\ref{tab:ablation_strategies}, are unequivocal. Med-REFL outperforms both the MCTS approach (even with a high number of simulations, `k=20') and the strong GPT-4.1-as-judge baseline. Notably, while increasing the number of rollouts for MCTS yields diminishing improvements, it never matches the performance of our method. This empirically validates our core hypothesis: the stable, holistic quality signal derived from our deterministic structural assessment is fundamentally more effective for teaching fine-grained reflection than either noisy statistical estimations or heuristic-based judgments. 
\begin{table}[htbp]
    \centering
    \small
    \captionsetup{skip=4pt}
    \caption{Ablation study comparing Med-REFL against alternative data generation strategies on MedQA-USMLE. Med-REFL's deterministic, action-focused evaluation provides a superior learning signal.}
    \label{tab:ablation_strategies}
    
        \begin{tabular}{llc}
        \toprule
        \textbf{Method} & Config. & ACC (\%) \\
        \midrule
        Base Model & - & 59.92 \\
        \midrule
        \multirow{3}{*}{MCTS-Rollouts} & $k=5$ & 60.76 \\
        & $k=10$ & 61.14 \\
        & $k=20$  & 61.21 \\
        \midrule
        LLM-as-Judge & GPT-4.1 & 60.59 \\
        \textbf{Ours} & - & \textbf{61.92} \\
        \bottomrule
        \end{tabular}
\end{table}
\paragraph{Detailed Experiment Settings}
To ensure a fair and rigorous comparison between Med-REFL and alternative data generation paradigms, we designed a controlled experimental environment where all methods operated on a unified set of reasoning steps. This appendix details the setup for this critical ablation study.

\paragraph{General Setup and Data Selection.}
All fine-tuning experiments for this study were conducted on the \textbf{Llama3.1-8B} base model. The data source was a pool of 1614 instances of incorrect reasoning nodes, denoted as `wrong nodes', sampled from the full set of error-containing paths generated during our initial tree-of-thoughts exploration on the MedQA-USMLE training set. To guarantee a fair comparison, all methods were tasked with evaluating the exact same candidate pool for each instance. This \textbf{unified candidate pool} consisted of the `wrong nodes' itself plus all of its sibling nodes from the original ToT tree. The shared ancestor path leading to these nodes is denoted as `history'.

\paragraph{Method A: Med-REFL (Control Group).}
Our control group follows the original Med-REFL methodology detailed in Section 3 of the main paper. For each candidate `error' node in the unified pool, we deterministically calculated its $v_{action}$ score with `next jump' nodes in other paths. The DPO preference pair was then constructed by contrasting the paths originating from the highest-scoring (`chosen') and lowest-scoring (`rejected') nodes.

\paragraph{Method B: MCTS-Multiple Rollouts (Experimental Group).}
This method was designed to be completely decoupled from the global structural information of the ToT. Same as method A, method B\&C also include the `reflection' in DPO pair construction.

\textbf{Scoring:} For each candidate node, we performed $k$ independent, random rollouts, with $k \in \{5, 10, 20\}$. The score $p$ for each node was calculated as $p = (N_{correct} - N_{incorrect}) / k$, where $N$ is the number of correct or incorrect rollouts.

\textbf{DPO Pair Construction:} We identified the nodes with the highest and lowest scores, $p_{max}$ and $p_{min}$. The `chosen' trajectory for the DPO pair was constructed from a randomly selected successful rollout initiated from the highest-scoring node. The `rejected' trajectory was constructed from a randomly selected failed rollout initiated from the lowest-scoring node. If a selected node had no successful (or failed) rollouts, the DPO pair for that data point was discarded.

\paragraph{Method C: LLM-as-Judge (Experimental Group).}
This method relied on the external, outcome-agnostic judgment of GPT-4.1.
\textbf{Scoring:} For each candidate node, we prompted GPT-4.1 to assign a quality score, $Score_{judge}$, on a scale of 1 to 5 based on its perceived medical accuracy and logical coherence, without any knowledge of the final answer.

\textbf{DPO Pair Construction:} The nodes with the highest and lowest judge scores were identified. A new, complete reasoning path was then generated starting from each of these two nodes to form the final `chosen'and `rejected' trajectories for the DPO pair.

\begin{figure}[h]
    \centering
    \begin{promptbox}[title={Prompt for MCTS Rollouts (performing a single rollout)}]
        \small 
        
        You are a medical expert completing a reasoning process for a USMLE-style question. You are given the original \textbf{QUESTION}, its \textbf{OPTIONS}, and a fixed reasoning \textbf{PATH} to follow.
        
        \vspace{0.5em}
        \textbf{\#\#\# Your Task}
        
        Your task is to continue this reasoning \textbf{PATH} to its logical conclusion and select the single best answer from the \textbf{OPTIONS}.
        
        \vspace{0.5em}
        \textbf{\#\#\# Output Format}
        
        Think step-by-step. At the very end of your entire response, you \textbf{MUST} state the final answer in the following exact format on a new line:
        
        \begin{tcolorbox}[colback=white, colframe=gray!30, boxrule=0.5pt, sharp corners]
        Final Answer: [OPTION]
        \end{tcolorbox}

        \vspace{0.5em}
        \textbf{\#\#\# Inputs}
        
        \begin{tcolorbox}[colback=white, colframe=gray!30, boxrule=0.5pt, sharp corners]
        QUESTION:\\
        \{question\}
        
        \vspace{0.5em}
        PATH:\\
        \{history\}\\
        \{next\_step\}
        \end{tcolorbox}
        
    \end{promptbox}
    \caption{The prompt used for single MCTS rollouts. Variables in curly braces are placeholders.}
    \label{fig:prompt_mcts}
\end{figure}

\begin{figure}[h]
    \centering
    \begin{promptbox}[title={Prompt for LLM-as-Judge Step Evaluation}]
        \small 
        
        You are an expert medical educator and a master of logical reasoning, tasked with evaluating a single step in a student's problem-solving process for a USMLE question.

        \textbf{\#\#\# Instructions}
        
        Evaluate the \textbf{CURRENT\_STEP} on a scale of 1 to 5 based on medical accuracy, logical coherence, and strategic value.
        
        \vspace{0.5em}
        \textbf{\#\#\# Scoring Scale}
        
        \begin{itemize} \setlength{\itemsep}{0pt} \setlength{\parskip}{0pt} \setlength{\parsep}{0pt}
            \item[\textbf{-}] \textbf{1 (Flawed/Harmful):} Medically incorrect or logically fallacious.
            \item[\textbf{-}] \textbf{2 (Weak):} Correct but unhelpful, irrelevant, or shows poor prioritization.
            \item[\textbf{-}] \textbf{3 (Acceptable):} Logical and correct, but generic or suboptimal.
            \item[\textbf{-}] \textbf{4 (Strong):} Logical, accurate, and strategically advances the reasoning.
            \item[\textbf{-}] \textbf{5 (Excellent):} Strong, and also demonstrates exceptional insight.
        \end{itemize}
        
        \vspace{0.5em}
        \textbf{\#\#\# Output Format}
        
        Your output \textbf{must} be a single, valid JSON object with two keys: \code{reasoning} and \code{score}.
        
        \vspace{0.5em}
        
        \begin{tcolorbox}[colback=white, colframe=gray!30, boxrule=0.5pt, sharp corners, left=2pt]
        \{ \\
        \ \ ``reasoning": ``Your brief explanation here...", \\
        \ \ ``score": ``your integer score from 1 to 5" \\
        \}
        \end{tcolorbox}
        
    \end{promptbox}
    \caption{The evaluation prompt used for the LLM-as-a-Judge mechanism.}
    \label{fig:judge_prompt}
\end{figure}

\subsection{\revise{Sensitivity and Robustness Analysis: A Reasoning Boundary Perspective}}
\label{appendix:sensitivity}

\revise{In this section, we analyze the sensitivity of our framework to key hyperparameters, specifically sampling parameters (seeds, temperature) and the action value weights ($\lambda$). We propose to view these parameters through the lens of the model's reasoning boundary~\citep{chen2024unlocking}.}

\revise{Since Med-REFL relies on learning the transition from erroneous states to correct states (i.e., self-correction), effective training data can only be generated from problems that lie within the model's reasoning boundary: the problem must be difficult enough to induce errors, yet solvable enough for the model to eventually find at least one correct path via search. Hyperparameters essentially determine the extent of this boundary explored during data generation.}

\subsubsection{\revise{Stability Across Sampling Parameters}}
\revise{The randomness in trajectory generation, controlled by random seeds and temperature ($T$), theoretically affects the specific paths explored in a single run. However, in the context of our ToT approach, this sensitivity is minimal. ToT functions similarly to a high $Pass@K$ sampling strategy, where multiple rollouts are aggregated. }

\revise{As long as the sampling parameters allow for sufficient diversity (e.g., $T > 0$), the probability of discovering a correct path for a problem within the model's reasoning boundary remains stable. Therefore, variations in seeds or slight fluctuations in temperature primarily shift \textit{which} valid path is found first, rather than \textit{whether} a path is found, ensuring the robustness of our method against stochastic noise.}

\subsubsection{\revise{Sensitivity of Action Value Weights ($\lambda$)}}
\label{sec:lambda_sensitivity}
\revise{The action value $v_{act}$, defined in Equation ~\ref{eq:action_value}, guides the preference learning by weighting three components: step value ($\lambda_1$), solution value ($\lambda_2$), and remaining value ($\lambda_3$). We interpret our chosen ratio of $\lambda_1:\lambda_2:\lambda_3 = 2:1:2$ from both qualitative and quantitative perspectives.}

\paragraph{\revise{Qualitative Analysis.}}
\revise{In the context of mid-reasoning reflection, the model evaluates a transition between nodes rather than a finalized solution.
\begin{itemize}
    \item We assign higher weights to \textbf{Step Value ($\lambda_1$)} and \textbf{Remaining Value ($\lambda_3$)} because the primary goal of reflection is to immediately correct a local error (Step Value) and ensure the corrected path has high potential to lead to a solution (Remaining Value).
    \item \textbf{Solution Value ($\lambda_2$)} is weighted lower because it represents a global outcome that may be distant and less discriminative for intermediate corrective actions.
\end{itemize}}

\begin{figure}[h]
    \centering
    \includegraphics[width=1\linewidth]{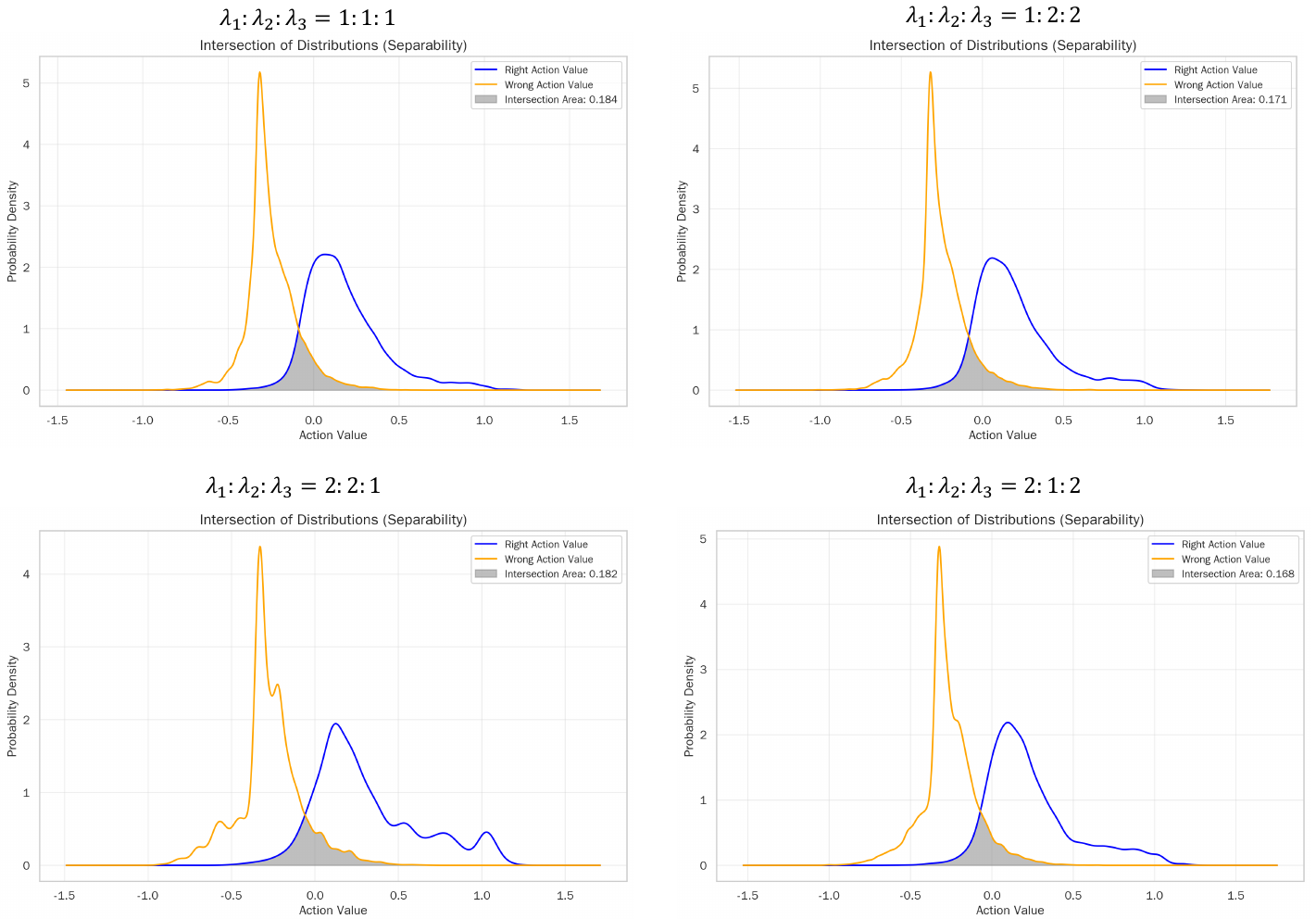}
    \caption{\revise{\textbf{Separability Analysis of Action Value Distributions under Different $\lambda$ Settings.} We plotted the probability density functions of action values for `Right Actions' (Blue) and `Wrong Actions' (Orange). The intersection area represents the confusion region. Our chosen setting of $\lambda_1:\lambda_2:\lambda_3 = 2:1:2$ achieves the minimum intersection area (0.168), indicating maximum separability and the strongest signal for distinguishing effective reflections.}}
    \label{fig:lambda_separability}
\end{figure}

\paragraph{\revise{Quantitative Separability Analysis.}}
\revise{To validate this empirically, we analyzed the discriminative power of different $\lambda$ configurations. We plotted the distribution of calculated action values for known `effective reflections' (right action) versus `ineffective reflections' (wrong action) and calculated the intersection area between the curves (Figure \ref{fig:lambda_separability}). A smaller intersection area implies better separability and a cleaner learning signal for the DPO process.}
\revise{\begin{itemize}
    \item $\lambda(1:1:1) \rightarrow \text{Intersection Area: } 0.184$
    \item $\lambda(1:2:2) \rightarrow \text{Intersection Area: } 0.171$
    \item $\lambda(2:2:1) \rightarrow \text{Intersection Area: } 0.182$
    \item \textbf{$\lambda(2:1:2) \rightarrow \text{Intersection Area: } 0.168$ (Optimal)}
\end{itemize}}
\revise{The results demonstrate that the 2:1:2 setting maximizes the distinction between correct and incorrect reflective actions, thereby providing the most robust supervision signal.}

\subsection{Generalization to Multimodal Reasoning}
\label{apdx:vlm_extension}

To further evaluate the cross-modal transferability of our framework, we extended the Med-REFL pipeline to Vision-Language Models (VLMs). 
We selected the Qwen3-VL-Instruct 8B as our base model and utilized the SLAKE~\citep{slake} training set as the seed corpus for trajectory generation. Following our core methodology, we adapted the ToT exploration and deterministic structural assessment to construct a fine-grained visual reasoning reflection dataset. As shown in Table~\ref{tab:vlm_results}, the model fine-tuned with our framework (denoted as +Med-REFL) demonstrates consistent improvements across three major medical VQA benchmarks. Notably, we observed a significant accuracy gain of +3.33\% on the VQA\_RAD\citep{vqa_rad} and 0.14\% on OmnimedVQA\citep{hu2024omnimedvqa}. These results suggest that Med-REFL instills a transferable self-correction mechanism that mitigates visual-reasoning misalignment, paving the way for more reliable multimodal AI assistants in high-stakes clinical domains.

\begin{table}[htbp]
\centering
\caption{Performance evaluation on Medical Visual Question Answering (VQA) benchmarks. All results represent Accuracy (\%).}
\label{tab:vlm_results}
\resizebox{\linewidth}{!}{
\begin{tabular}{lccc}
\toprule
\textbf{Model} & \textbf{SLAKE} & \textbf{VQA\_RAD} & \textbf{OmnimedVQA} \\
\midrule
Qwen3-VL-Instruct 8B & 66.16 & 57.20 & 74.51 \\
\textbf{Qwen3-VL-Instruct 8B + Med-REFL} & \textbf{67.11 (+0.95)} & \textbf{60.53 (+3.33)} & \textbf{74.65 (+0.14)} \\
\bottomrule
\end{tabular}
}
\end{table}
\section{Case Study}
\label{apdx:case_study}
Except for Figure~\ref{fig:casestudy} (Q1), to provide a more intuitive and granular understanding of how the Med-REFL framework enhances the complex medical reasoning and self-correction capabilities of Large Language Models (LLMs), this appendix presents two detailed case studies: dpo-pairs case study and huatuo-o1's thinking pattern change after being trained with Med-REFL.

\subsection{Med-REFL Refines the Reasoning of Huatuo-o1}
\begin{figure}[htbp]
    \centering
    \includegraphics[width=1\linewidth]{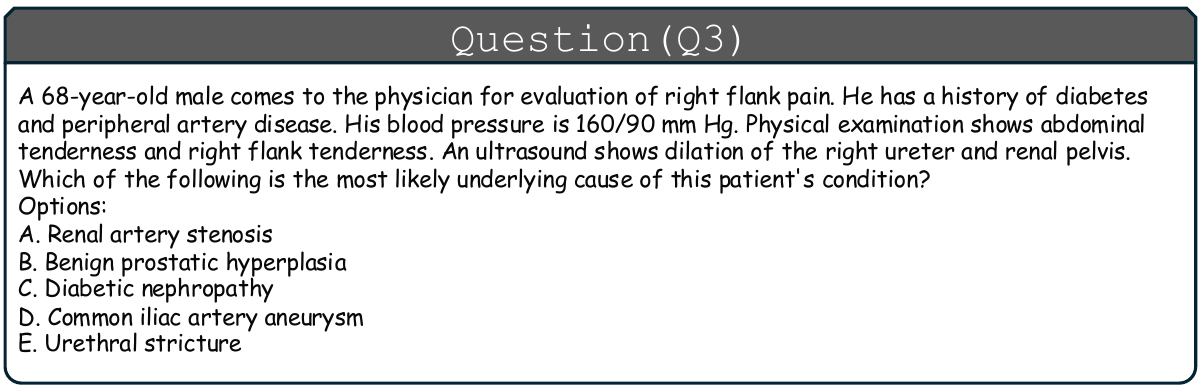}
    \caption{The clinical scenario for Question 3, a complex case of unilateral hydronephrosis in a patient with significant vascular comorbidities.}
    \label{fig:q3_question}
\end{figure}

This case study shows a shift from reasoning based on `statistical prevalence' to a more precise diagnostic process based on `patient-specific evidence' for the case presented in Figure \ref{fig:q3_question}.

\begin{figure}[htbp]
    \centering
    \includegraphics[width=1\linewidth]{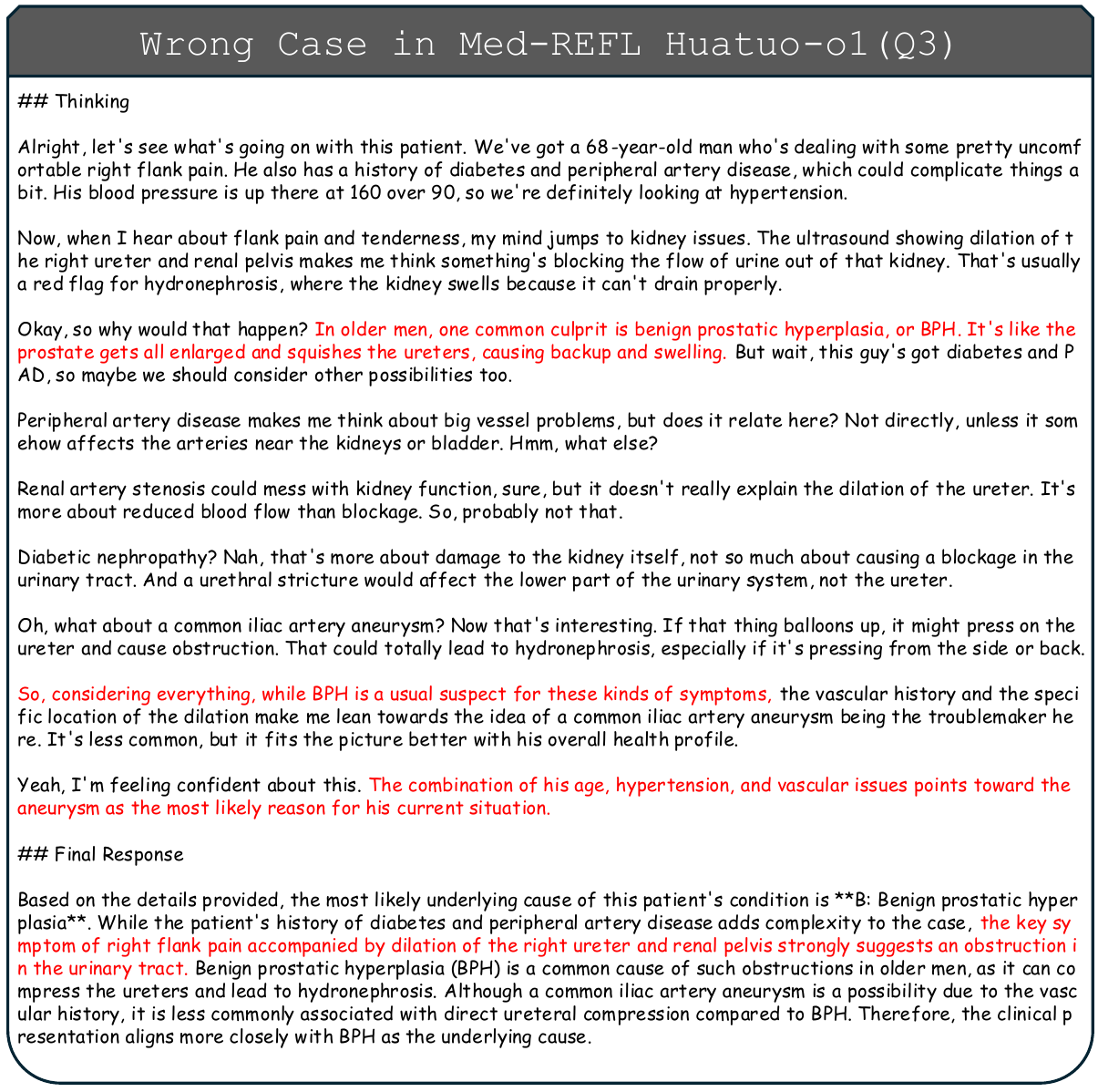}
    \caption{The reasoning chain of the baseline Huatuo-o1 model for Q3. While identifying the correct differential, it defaults to the statistically common but less fitting diagnosis of BPH.}
    \label{fig:q3_wrong_cot}
\end{figure}

\begin{figure}[htbp]
    \centering
    \includegraphics[width=1\linewidth]{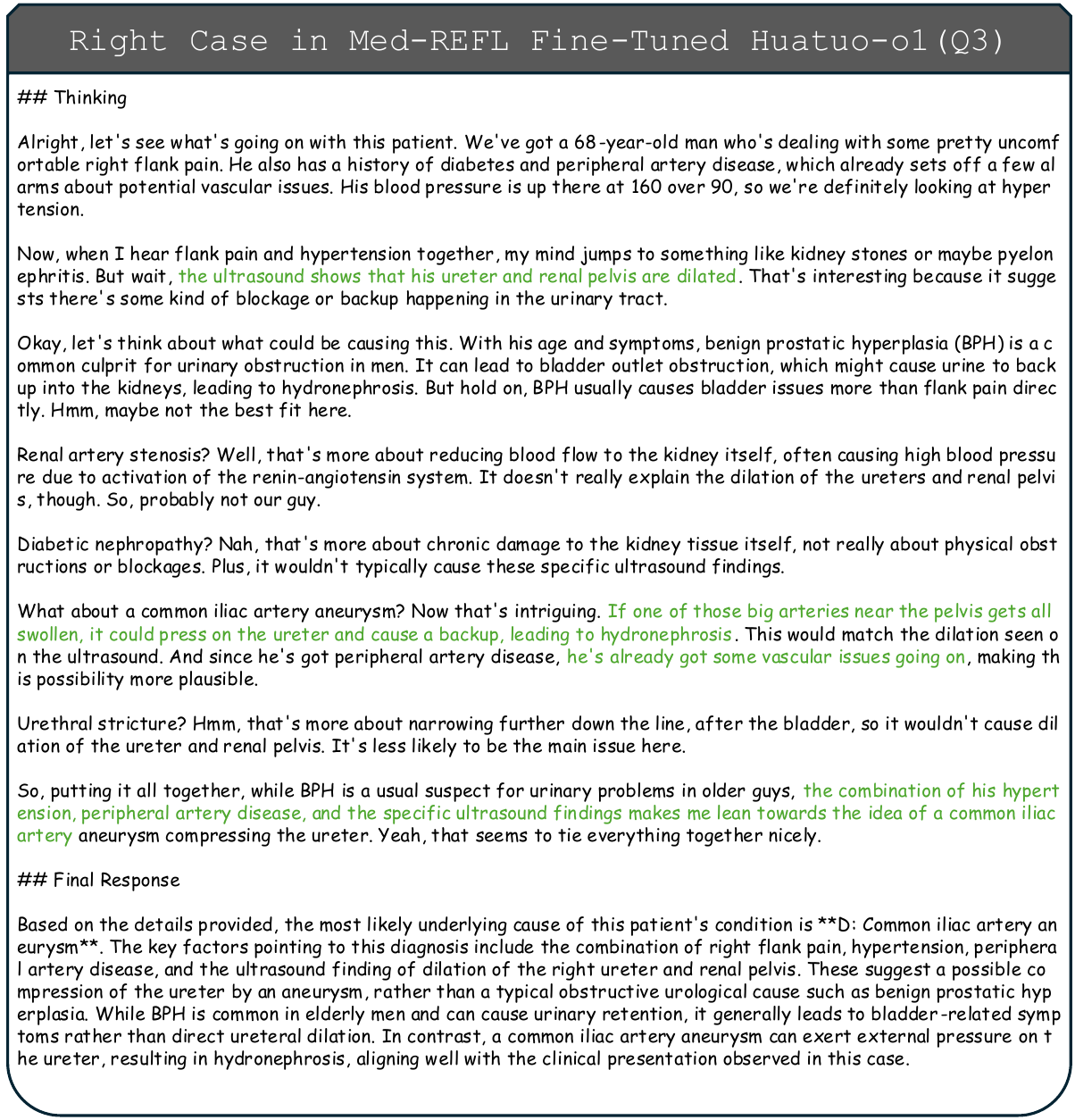}
    \caption{The refined reasoning chain of Huatuo-o1 after Med-REFL fine-tuning for Q3. The model demonstrates superior clinical acumen by critiquing the common diagnosis and correctly prioritizing patient-specific evidence.}
    \label{fig:q3_right_cot}
\end{figure}

\paragraph{Case Summary}
Med-REFL refines Huatuo-o1's reasoning capability, elevating it from a model that knows `what is common' to a clinical reasoner that understands `what is most likely, given this specific patient's evidence.' This demonstrates that the framework enhances the model's ability to weigh evidence appropriately, leading to more accurate and logically sound diagnostic decisions.

\subsection{DPO Pairs}
\begin{figure}[htbp]
    \centering
    \includegraphics[width=1\linewidth]{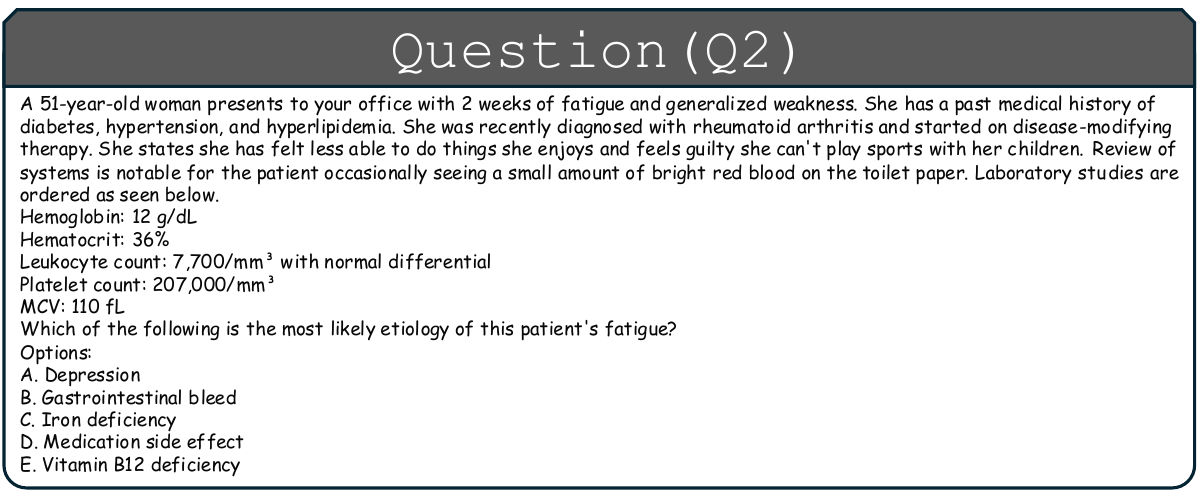}
    \caption{The clinical scenario for Question 2, presenting contradictory clues of macrocytic anemia and signs of a gastrointestinal bleed.}
    \label{fig:q2_question}
\end{figure}

A core mechanism of Med-REFL is the construction of high-quality preference data, consisting of a ‘Chosen’ path of effective reflection and a ‘Rejected’ path of ineffective reflection. This case study presents a typical DPO pair used to train the model how to navigate a complex case with contradictory clinical information (Figure \ref{fig:q2_question}).

\begin{figure}[htbp]
    \centering
    \includegraphics[width=0.8\linewidth]{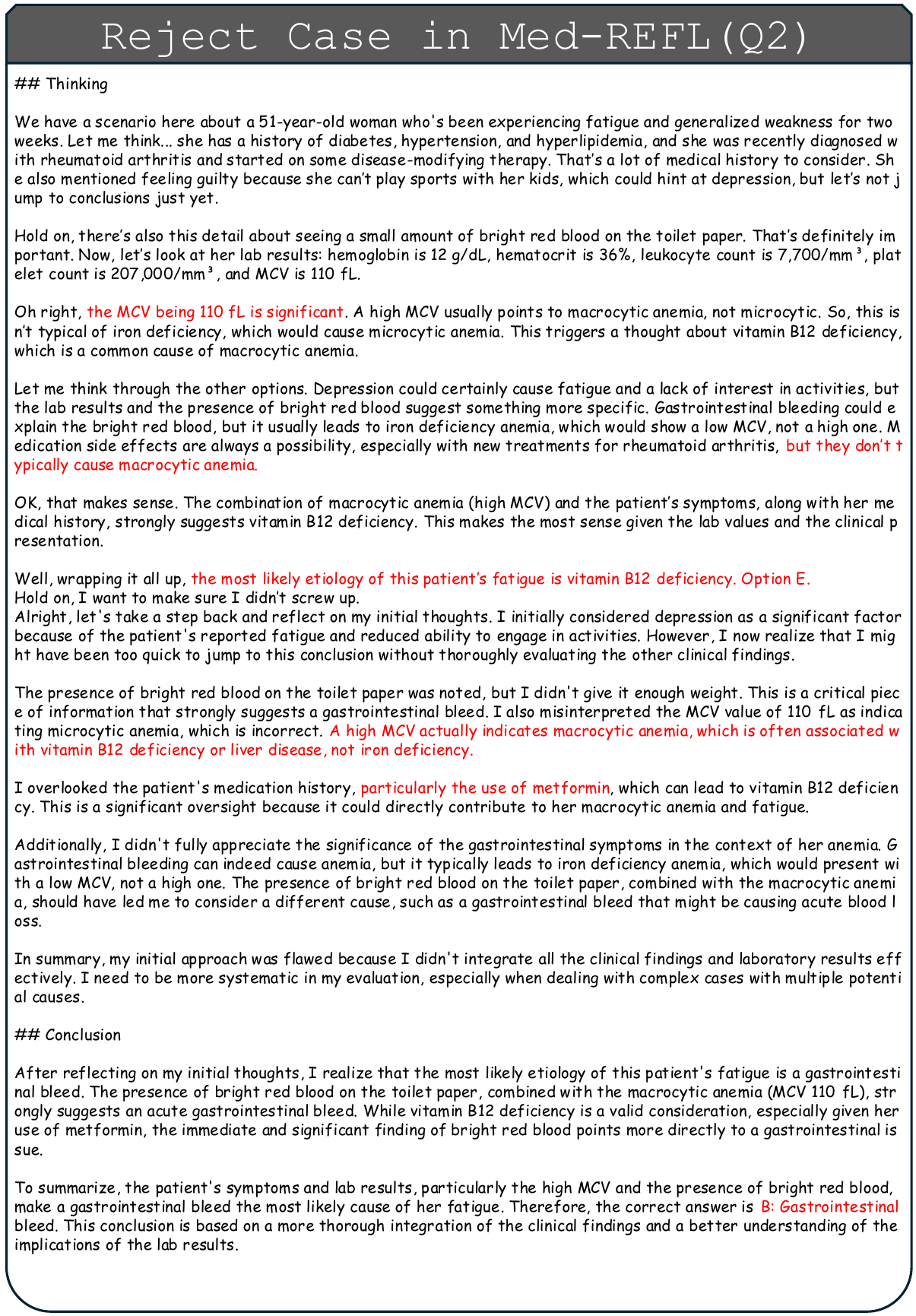}
    \caption{The ‘Rejected’ reasoning path for Q2. The model's reflection fails to reconcile the conflict between a GI bleed and macrocytic anemia, leading to an incorrect conclusion.}
    \label{fig:q2_rejected_cot}
\end{figure}

\paragraph{The ‘Rejected’ Path (Ineffective Reflection)}
This path demonstrates a confused reasoning process that fails to integrate all clinical clues (Figure \ref{fig:q2_rejected_cot}).

\begin{figure}[htbp]
    \centering
    \includegraphics[width=0.8\linewidth]{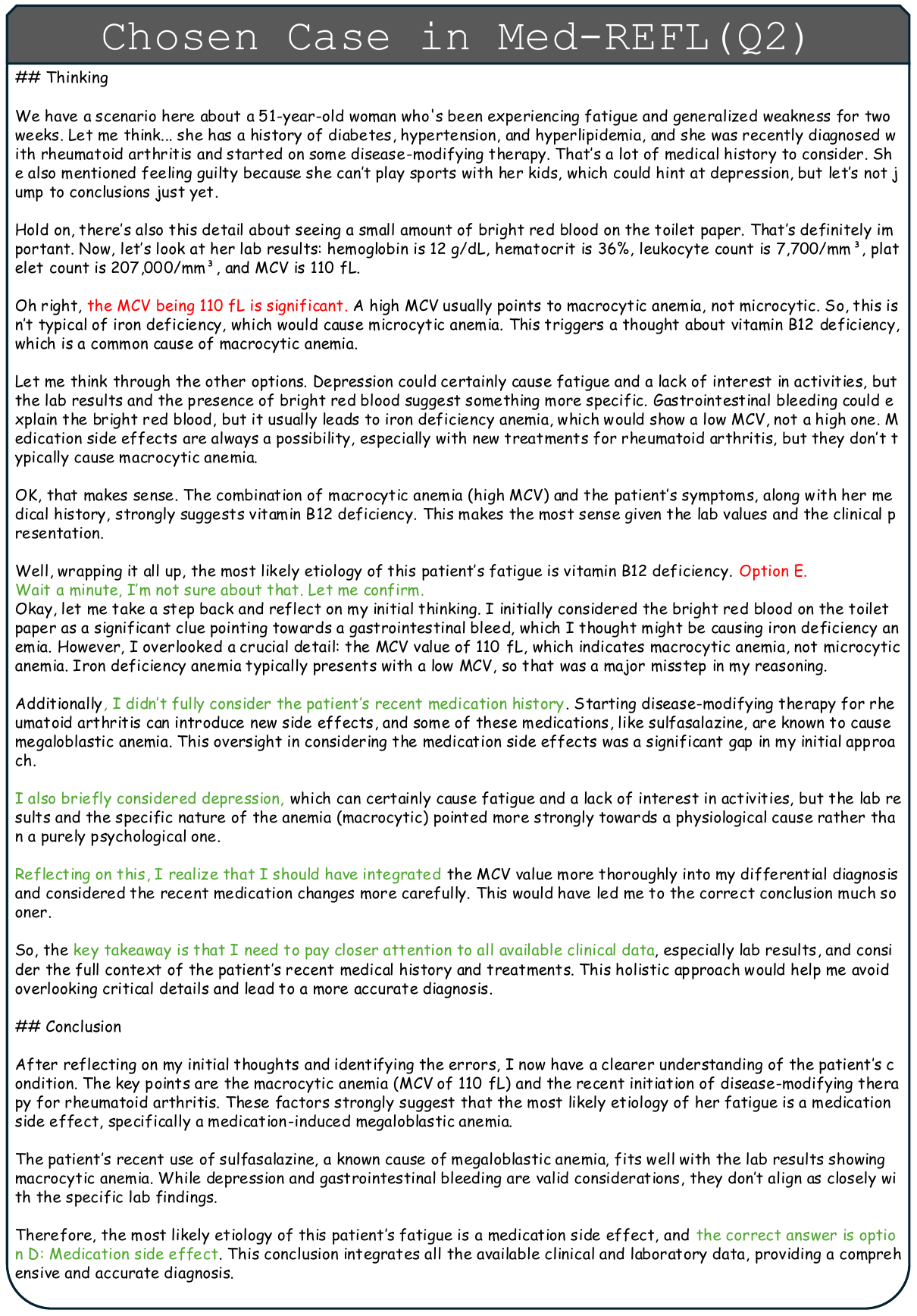}
    \caption{The ‘Chosen’ reasoning path for Q2, which forms the positive example in the DPO pair. The model correctly identifies the central contradiction and integrates the medication history to arrive at the correct diagnosis.}
    \label{fig:q2_chosen_cot}
\end{figure}

\paragraph{The ‘Chosen’ Path (Effective Reflection)}
This path demonstrates a superior reasoning process that logically integrates all available data (Figure \ref{fig:q2_chosen_cot}).

\paragraph{Case Summary}
This DPO pair explicitly teaches the model that when faced with conflicting clinical clues, superior reasoning involves finding a diagnosis that explains \textbf{all} key data points, rather than cherry-picking one and ignoring others. Through such contrastive learning, the model develops a preference for effective reflection that resolves clinical contradictions and integrates complex information holistically.

\end{document}